\documentclass{article} 
\usepackage{iclr2027_conference,times}

\usepackage{amsmath,amsfonts,bm}

\def\eqref#1{equation~\ref{#1}}

\def\1{\bm{1}}

\DeclareMathAlphabet{\mathsfit}{\encodingdefault}{\sfdefault}{m}{sl}
\SetMathAlphabet{\mathsfit}{bold}{\encodingdefault}{\sfdefault}{bx}{n}

\usepackage{hyperref}
\usepackage{url}
\usepackage{amsmath}
\usepackage{amssymb}
\usepackage{booktabs}
\usepackage{graphicx}

\usepackage{pifont}
\usepackage{placeins}
\usepackage{enumitem}

\usepackage{array}
\usepackage{longtable}
\usepackage{multirow}
\usepackage{wrapfig}
\usepackage{xcolor}
\usepackage[table]{xcolor}
\usepackage{makecell}

\definecolor{gainGreen}{HTML}{01AF50}
\definecolor{dropRed}{HTML}{E05252}
\newcommand{\mygreen}[1]{\textcolor{gainGreen}{#1}}
\newcommand{\myred}[1]{\textcolor{dropRed}{#1}}

\newcommand{\cmark}{\textcolor{green}{\ding{51}}}
\newcommand{\xmark}{\textcolor{red}{\ding{55}}}

\definecolor{bestgreen}{RGB}{226, 246, 226}
\definecolor{secondgreen}{RGB}{242, 251, 242}

\usepackage{booktabs}
\usepackage{tabularx}

\title{Acting in Meters: Learning Metric Interactions for Precise Robotic Manipulation}

\author{
  \bfseries Lijie Wang \dag, Zheng Lu \dag, Yiming Wang \dag, Heyang Yu, Kenghou Hoi, Bowen Hu, \\
  \multicolumn{1}{c}{%
    \bfseries Di Cui, Tianyu Xin, Haoran Liao, Wanqi Zhong, Xingjie Fan, Yizhao Xu,
  } \\
  \multicolumn{1}{c}{%
    \bfseries Ziliang Wang, Fei Gao, Yiming Li%
    \thanks{\textbf{Corresponding author}: Fei Gao and Yiming Li. \dag \textbf{Equal contribution}. Lijie Wang, Kenghou Hoi, Bowen Hu, and Fei Gao are with Zhejiang University; Zheng Lu and Yizhao Xu are with Peking University; Heyang Yu, Tianyu Xin, Wanqi Zhong, Ziliang Wang, Xingjie Fan and Yiming Li are with Tsinghua University; Yiming Wang is with Shanghai Jiao Tong University;
    Di Cui is with Nankai University;
    Haoran Liao is with Sun Yat-sen University. \textit{\textbf{Email}: lijiewang@zju.edu.cn}} 
  }
}

\iclrfinalcopy 
\begin{document}

\maketitle

\begin{abstract}
Vision-Language-Action (VLA) models and World-Action Models (WAMs) have increasingly advanced general language-conditioned robotic manipulation, yet often
leave metric relations among actions, manipulated objects, and scene geometry
implicit. Human manipulation combines semantic understanding of
task-relevant objects with spatial feedback that guides hand motion
relative to objects and their surroundings. Inspired by this, we introduce a metric interaction framework that
models object-level and scene-level interactions in physical Cartesian space
at a shared metric scale.
At the object level, Interaction-Centric Tokens (ICTs) explicitly
represent end-effector pose trajectories relative to manipulated
objects and are jointly denoised with actions, providing physically
grounded interaction supervision.
At the scene level, the Metric Action Interaction Field (MAIF)
uses action and ICT queries to attend to metric scene point-cloud
features and learns geometry-conditioned action corrections.
Through two-stage adaptation, our framework improves diverse VLA
and WAM baselines with a small number of additional parameters and training steps.
Experiments demonstrate average success-rate gains of 0.80 and 3.59
percentage points on LIBERO and RoboTwin~2.0, respectively,
alongside gains of 6.80 percentage points on real-world tasks
and 7.45 percentage points on their out-of-distribution variants.
\end{abstract}

\section{Introduction}
\begin{figure}[!h]
  \centering
  \includegraphics[width=1.0\linewidth]{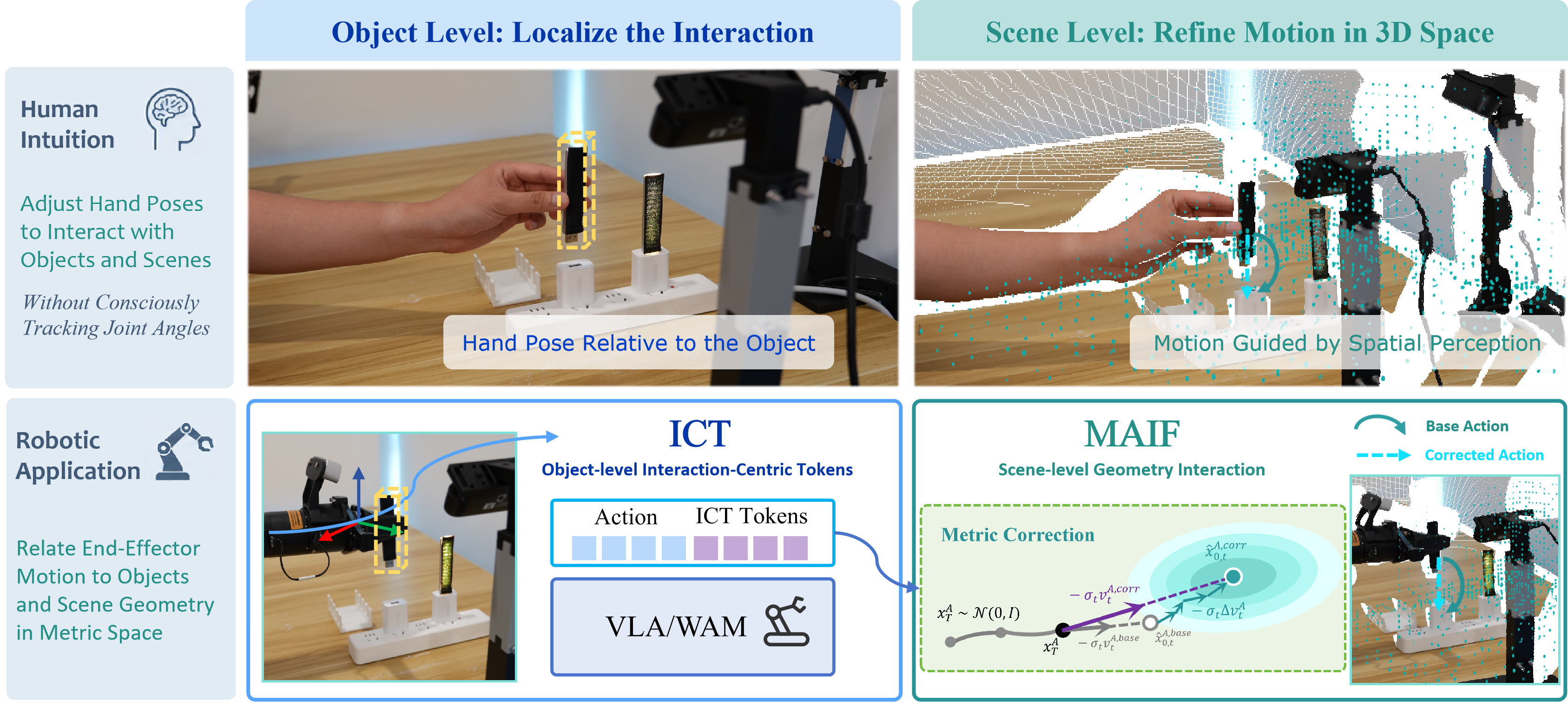}
\caption{\textbf{From human intuition to metric interaction.}
\textbf{Top:} Humans use spatial perception to adjust hand poses
relative to objects and surroundings.
\textbf{Bottom:} We adapt robotic manipulation policies to generate and refine
actions by explicitly modeling end-effector interactions with
objects and scene geometry in Cartesian space at a shared
metric scale.}
  \label{fig:headimg}
\end{figure}

Vision-Language-Action (VLA) models have evolved from discretized action-token prediction~\cite{rt-2,openvla} to continuous action-chunk generation with diffusion or flow-matching experts~\cite{pi0,pi0.5,pi0.7}, enabling increasingly general language-conditioned manipulation across tasks and embodiments. World-Action Models (WAMs) further incorporate predictive world modeling, either recovering actions from imagined futures~\cite{unipi,robodreamer} or jointly modeling future observations and actions for end-to-end control~\cite{uwm,dreamzero,lingbot-va,fast-wam}. More recently, geometry-aware policies introduce depth, point clouds or 3D motion priors to improve spatially precise manipulation~\cite{spatialvla,spatialforcing,X-WAM,wam4d,meco-wam,lamp,mu0,pointaction}.

Despite these advances, VLAs and WAMs operating in joint space typically learn metric spatial relations among end--effectors, objects, and the surroundings implicitly from action supervision. 
Geometry-aware approaches strengthen spatial modeling in policies
by incorporating depth \cite{X-WAM}, point maps \cite{pointworld}, or features from foundation
models \cite{spatialvla,falcon}; introducing auxiliary objectives for geometric prediction
and representation alignment \cite{meco-wam, spatialforcing}; or conditioning action generation
on predicted motion intermediates such as 3D scene flow. \cite{pointaction, mu0, lamp}
While these approaches provide spatial and motion cues, stronger geometric perception does not necessarily translate into explicit metric interaction modeling. In particular, enriching geometric representations does not ensure that object-relative end-effector states are supervised or that candidate trajectories are grounded in surrounding scene with a consistent coordinate and physical scale. The challenge is therefore not merely to perceive the scene in 3D, but to explicitly model the metric relations that connect robotic actions with objects and the environment.

Human manipulation provides a task-space perspective on this
challenge as illustrated in
Fig.~\ref{fig:headimg}. When grasping a cup, semantic recognition identifies
the target, while successful execution requires adjusting the
hand's pose relative to the cup and its
surroundings through spatial perception guidance. During this process, people
typically focus on the hand's pose relative to the target rather
than consciously tracking individual shoulder or elbow joint angles.
This task-space perspective suggests that robotic policies should
explicitly relate end-effector motion to manipulated objects and
surrounding geometry. Modeling these interactions in Cartesian
space at a shared metric scale provides a direct interface between
semantic understanding and physically grounded action generation.

We realize this idea at two complementary levels.
At the object level, we adapt HumanEgo's Interaction-Centric Tokens
(ICTs)~\citep{wang2026humanego} to represent end-effector pose
trajectories relative to manipulated objects. These tokens are
jointly denoised with actions, providing physically grounded
interaction supervision.
At the scene level, we propose the Metric Action Interaction Field
(MAIF), which relates decoded end-effector trajectories to metric
scene geometry. During training, its action and ICT queries
cross-attend to scene features encoded from 3D point coordinates and
surface normals to learn geometry-conditioned flow corrections.
We first adapt the action expert with ICT, then freeze the adapted
policy to train MAIF. The framework preserves each backbone's native
action representation and sampling schedule.
Our contributions are threefold: \textbf{(1)} We introduce an object-level metric interaction mechanism that encodes the relative poses between end effectors and objects using Interaction-Centric Tokens, providing the policy backbone with physically grounded interaction supervision. \textbf{(2)} We propose the Metric Action Interaction Field,
a scene-level geometry-aware residual adapter that relates decoded
end-effector trajectories to metric scene geometry and refines
action predictions while preserving the backbone's native sampling
schedule. \textbf{(3)} We validate our framework on both WAM and VLA backbones across simulation benchmarks and real-world manipulation tasks, demonstrating improvements in task success rate and generalization.

\section{Related Works}


\subsection{Vision-Language-Action (VLA)}

VLA models use pretrained vision-language models (VLMs) for perception and reasoning in robot control.
MiMo-Embodied~\citep{hao2025mimoembodied} develops spatial understanding and task planning, while Causal Planner~\citep{lu2026tokenpredictors} targets causal reasoning.
For action generation, LayerRoute~\citep{lu2026layerroute} uses the current action representation to combine features from different VLM layers.
RT-2 transfers web-scale semantics through action tokenization~\citep{rt-2}, OpenVLA provides an open generalist policy~\citep{openvla}, and $\pi_0$ introduces a flow-matching expert for continuous action chunks~\citep{pi0}.
Models including $\pi_{0.5}$, GR00T N1, RDT-1B, and SmolVLA extend to open-world, humanoid, bimanual, and low-cost settings~\citep{pi0.5,gr00tn1,rdt-1b,smolvla}.
$\pi_{0.7}$ exploits multimodal context~\citep{pi0.7}, while LingBot-VLA 2.0 combines large-scale cross-embodiment pretraining with video and depth prediction~\citep{lingbotvla2}.


\subsection{World-Action Model (WAM)}

Early imagine-then-act systems, including UniPi, RoboDreamer, GR-2, and DreamGen, generate visual futures and recover actions via inverse dynamics or downstream control~\cite{unipi,robodreamer,gr-2,dreamgen}, while UVA, UWM, and Cosmos Policy couple video and action generation more tightly~\cite{uva,uwm,cosmospolicy}. Native models LingBot-VA and LingBot-VA 2.0 learn causal or joint video-action representations for generalizable control~\cite{lingbot-va,lingbot-va-2.0}; GigaWorld-Policy and GigaWorld-Policy-0.5 emphasize efficient action-centered prediction~\cite{gigaworld-policy,gigaworld-policy-0.5}. Fast-WAM benefits from video co-training without test-time imagination~\cite{fast-wam}, while WAM4D incorporates 4D geometric priors through spatial register tokens~\cite{wam4d}. 

\subsection{Geometry Modeling in Embodied Models}

LiDAR-VGGT \cite{lidar-vggt} utilizes LiDAR to assist gerometry foudation model VGGT \cite{vggt}.
SpatialVLA uses Ego3D encoding and adaptive action
grids~\cite{spatialvla}; Spatial Forcing aligns VLA features with a
pretrained 3D model~\cite{spatialforcing}; and GAM repurposes a geometric
foundation model for perception, prediction, and action
decoding~\cite{Gam}. World models capture 4D dynamics through geometry
prediction or distillation: TesserAct predicts RGB, depth, and
normals~\cite{tesseract}, while X-WAM jointly generates multi-view
RGB-D futures and actions~\cite{X-WAM}. GEM-4D distills dense
correspondences into video generation~\cite{gem-4d}, and EgoGenesis
stabilizes egocentric rollouts with anchored projective memory and
3D action encoding~\cite{Egogenesis}. GeoSem-WAM augments RGB prediction
with geometry and semantics~\cite{geosem-wam}, while WAM4D and MECo-WAM
transfer 4D priors through registers or auxiliary
experts for efficient deployment~\cite{wam4d,meco-wam}.

\section{Method}
\begin{figure}[!h]
  \centering
  \includegraphics[width=1.0\linewidth]{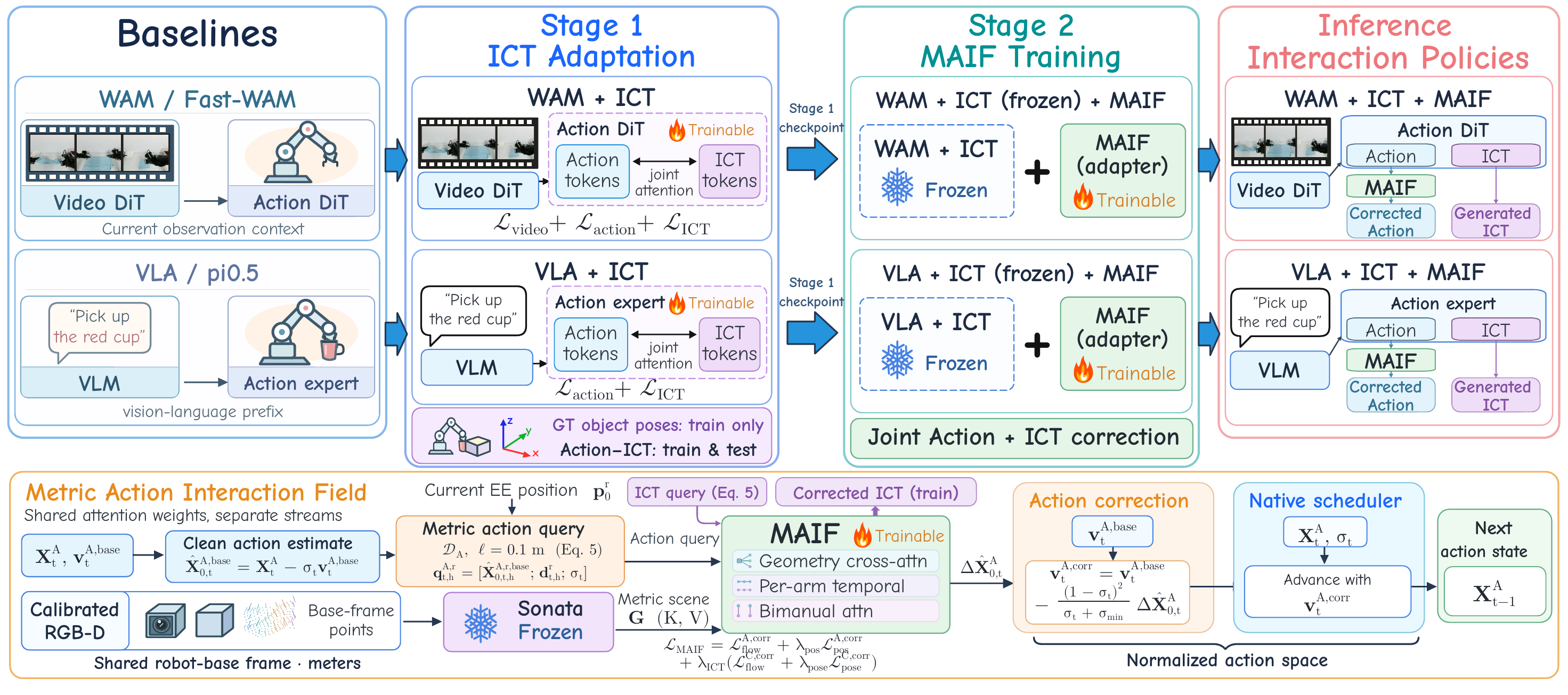}
  \caption{\textbf{Framework Overview.}
  \textbf{Top:} Stage~1 jointly trains action and ICT prediction within
  each backbone's action expert. Stage~2 freezes the adapted policy and
  trains MAIF for joint action and ICT refinement.
  \textbf{Bottom:} MAIF combines cross-attention to metric scene features
  with temporal and bimanual attention to predict action-flow corrections.
  At inference, MAIF applies gated corrections only to action velocities,
  preserving the backbone's sampling schedule.}
  \label{fig:pipeline}
\end{figure}

Fig.~\ref{fig:pipeline} presents our two-stage framework for VLA and
WAM backbones. Stage~1 introduces action-aligned, object-relative pose
trajectories into the action expert through ICT. Stage~2 freezes the
adapted policy and trains MAIF to refine its predictions in native
flow space using scene geometry. Expert actions and
synchronized end-effector poses are obtained directly from simulation
trajectories and recorded during real-world teleoperation.

\subsection{Interaction-Centric Token (ICT)}
\label{sec:ict}

\paragraph{Metric interaction trajectory.}
Let $H$ denote the horizon length, $h\in\{1,\ldots,H\}$ index its
steps, and $r\in\{L,R\}$ index the arms; $o_h$ denotes the manipulated
object at step $h$. Object identities and poses are obtained from
simulator states or real-world annotations. A superscript $*$ denotes
a training target. The synchronized poses
$T_{o_h}^{W,*},T_{e_h^r}^{W,*}\in SE(3)$ map object and end-effector
coordinates into the world frame $W$. ICT encodes their relative pose as
\begin{equation}
\mathbf c_h^r
=\psi\!\left((T_{o_h}^{W,*})^{-1}T_{e_h^r}^{W,*}\right)
=[\mathbf p_h^{C,r,*};
  \operatorname{rot6d}(\mathbf R_h^{C,r,*})]
\in\mathbb R^9.
\label{eq:ict_encoding}
\end{equation}
Here $\mathbf p_h^{C,r,*}\in\mathbb R^3$ is the translation in meters
and $\mathbf R_h^{C,r,*}\in SO(3)$ is the rotation matrix, both relative
to the object frame; $C$ identifies the ICT stream. The mapping $\psi$
concatenates translation and $\operatorname{rot6d}(\mathbf R)$, which
stacks the first two rotation-matrix columns. Stacking
$[\mathbf c_h^L;\mathbf c_h^R]$ over $H$ steps gives
$\mathbf C_0\in\mathbb R^{H\times18}$. Ground-truth object poses are
used only for training supervision.

\paragraph{Action-side integration.}
We jointly denoise normalized action chunks
$\mathbf X_0^A=\mathbf A_0\in\mathbb R^{H\times D_a}$ and ICT targets
$\mathbf X_0^C=\mathcal N_C(\mathbf C_0)$, where $D_a$ is the native
action dimension per horizon step and $\mathcal N_C$ is a fixed,
invertible, elementwise affine normalization. Let $t$ index denoising
stages and $\sigma_t\in[0,1]$ denote the corresponding noise level.
During training, we sample a common noise level $\sigma_t$ for both
streams from the backbone's flow distribution. For $z\in\{A,C\}$
indexing action and ICT, respectively, the normalized noisy states
and flow targets are
\begin{equation}
\mathbf X_t^z
=(1-\sigma_t)\mathbf X_0^z+\sigma_t\boldsymbol\epsilon^z,
\qquad
\mathbf v^{z,*}=\boldsymbol\epsilon^z-\mathbf X_0^z.
\label{eq:ict_flow_path}
\end{equation}
The noise terms $\boldsymbol\epsilon^A$ and $\boldsymbol\epsilon^C$ are
independent standard Gaussians on valid coordinates; flow velocities
are defined with respect to the noise coordinate. The streams share
the action blocks with separate input projections and velocity heads.
Each horizon step has one token per stream; ICT tokens follow action
tokens with matching rotary positions and shared noise-level
conditioning. Invalid arm and padding coordinates are zeroed before
projection. Both streams attend to each other and the
observation--language context and are jointly denoised at inference.

\paragraph{Flow and relative-pose supervision.}
ICT receives flow supervision in normalized coordinates and pose
supervision in the object frame. Let $\mathcal V$ index valid
$(b,h,r)$ slots (sample, step, arm), excluding unavailable arms and
padded steps. In a batch, $\sigma_{t,b}$ denotes the noise level of
sample $b$. From the predicted velocity $\mathbf v_t^C$, we recover
a clean estimate and evaluate its decoded pose error:
\begin{equation}
\begin{aligned}
\hat{\mathbf X}_{0,t}^C
&=\mathbf X_t^C-\sigma_t\mathbf v_t^C, ~~~
(\hat{\mathbf P}_t^C,\hat{\mathbf R}_t^C) =\mathcal D_C(\hat{\mathbf X}_{0,t}^C),
\\
s_{t,bhr}
&=\frac{\|\hat{\mathbf p}_{t,bhr}^C-\mathbf p_{bhr}^{C,*}\|_2^2}{s_p^2}
+\frac{\|\hat{\mathbf R}_{t,bhr}^C-\mathbf R_{bhr}^{C,*}\|_F^2}{2s_\theta^2}.
\end{aligned}
\label{eq:ict_pose_discrepancy}
\end{equation}
Hats mark estimates of clean states and decoded poses;
$\hat{\mathbf P}_t^C$ and $\hat{\mathbf R}_t^C$ collect the per-slot
translations $\hat{\mathbf p}_{t,bhr}^C$ and rotation matrices
$\hat{\mathbf R}_{t,bhr}^C$. The fixed decoder $\mathcal D_C$ applies
$\mathcal N_C^{-1}$, extracts translation, and decodes rotation by
column-wise, right-handed Gram--Schmidt, using a fixed rotation in
$SO(3)$ as a fallback for degenerate codes. Fixed scales $s_p>0$
(meters) and $s_\theta>0$ (radians) normalize translation and chordal
rotation errors.

Flow matching uses all sampled noise levels, while pose supervision
uses only the low-noise subset
$\mathcal V_{\mathrm{geo}}
=\{(b,h,r)\in\mathcal V:0<\sigma_{t,b}\leq\sigma_{\mathrm{geo}}\}$,
with fixed $0<\sigma_{\mathrm{geo}}<1$:
\begin{equation}
\begin{aligned}
\mathcal L_{\mathrm{flow}}^C
&=\frac{1}{9|\mathcal V|}
\sum_{(b,h,r)\in\mathcal V}
w_{\mathcal B}(\sigma_{t,b})
\|\mathbf v_{t,bhr}^C-\mathbf v_{bhr}^{C,*}\|_2^2,~~
\mathcal L_{\mathrm{pose}}^C=\frac{1}{|\mathcal V_{\mathrm{geo}}|}
\sum_{(b,h,r)\in\mathcal V_{\mathrm{geo}}}
\rho_{\mathrm H}(s_{t,bhr}),
\\
\mathcal L_{\mathrm{ICT}}
&=\mathcal L_{\mathrm{flow}}^C
+\lambda_{\mathrm{pose}}\mathcal L_{\mathrm{pose}}^C.
\end{aligned}
\label{eq:ict_combined_loss}
\end{equation}
Here $\mathcal B$ identifies the backbone, $w_{\mathcal B}$ is its
native flow weight, and fixed $\lambda_{\mathrm{pose}}>0$ balances
the two terms. The robust penalty is $\rho_{\mathrm H}(s)=s$ for
$0\leq s\leq1$ and $2\sqrt{s}-1$ otherwise. Decoding for the ICT pose
loss is restricted to $\mathcal V_{\mathrm{geo}}$; each loss term is
zero when its averaging set is empty. Stage~1 adds
$\mathcal L_{\mathrm{ICT}}$ to the backbone's native training objective.

\subsection{Metric Action Interaction Field (MAIF)}
\label{sec:maif}

\paragraph{Metric interaction queries.}
For stream $z\in\{A,C\}$, the frozen ICT-adapted expert predicts
$\mathbf v_t^{z,\mathrm{base}}$, yielding the clean estimate
$\hat{\mathbf X}_{0,t}^{z,\mathrm{base}}
=\mathbf X_t^z-\sigma_t\mathbf v_t^{z,\mathrm{base}}$
under Eq.~\eqref{eq:ict_flow_path}. The superscript $\mathrm{base}$
identifies predictions from the frozen policy. A fixed action metric
decoder $\mathcal D_A$ maps clean action estimates to end-effector
positions $\hat{\mathbf p}_{t,h}^{A,r,\mathrm{base}}$ in the robot base
frame by undoing action normalization and applying forward kinematics. Depth observations are back-projected into a point cloud
in the robot base frame. A frozen Sonata~\cite{sonata} encoder
processes the 3D point coordinates and surface normals, without
RGB attributes. We concatenate its multi-scale features into scene
tokens $\mathbf G$; feature aggregation and attention implementation
details are provided in Appendix~\ref{appendix}. Point coordinates
are expressed in meters, without per-scene centering or rescaling.

We split the clean action and ICT predictions into arm--step slots
and construct paired query features:
\begin{equation}
\begin{aligned}
\mathbf q_{t,h}^{A,r}
&=[\hat{\mathbf X}_{0,t,h}^{A,r,\mathrm{base}};
\mathbf d_{t,h}^r;\sigma_t],~~~
\mathbf q_{t,h}^{C,r}=[\phi_C(\hat{\mathbf c}_{0,t,h}^{r,\mathrm{base}});
\mathbf d_{t,h}^r;\sigma_t].
\end{aligned}
\label{eq:maif_queries}
\end{equation}
Here $\mathbf d_{t,h}^r
=(\hat{\mathbf p}_{t,h}^{A,r,\mathrm{base}}-\mathbf p_0^r)/\ell$
is the predicted base-frame displacement from the current measured
end-effector position $\mathbf p_0^r$, with fixed
$\ell=0.1\,\mathrm m$. The ICT feature
$\hat{\mathbf c}_{0,t,h}^{r,\mathrm{base}}$ is the corresponding slot
of $\mathcal N_C^{-1}(\hat{\mathbf X}_{0,t}^{C,\mathrm{base}})$;
$\phi_C$ divides its translation by $\ell$ and retains the six rotation
coordinates. Thus, both queries include the same predicted motion
feature, while the ICT query additionally encodes the object-relative
interaction state. Its translation remains in the object's reference
frame.

Let $\mathbf q_t^z$ collect the arm-step features of stream $z$.
Stream-specific projections produce hidden query sequences of a
common width, $\mathbf Q_t^z=E_z(\mathbf q_t^z)$.
Within each stream, queries cross-attend to metric scene tokens
$\mathbf G$, which supply keys and values. Temporal attention
connects horizon steps within each arm, while bimanual attention
exchanges information between the two arms at matching steps.
The streams share attention parameters but are processed separately,
without action-ICT attention within MAIF. Output heads are
stream-specific. MAIF has about \textit{\textbf{200M}} trainable parameters.

\paragraph{Gated flow correction.}
All base predictions supplied to MAIF are detached. Zero-initialized
output heads predict correction proposals in normalized clean space,
which are converted to gated velocity corrections:
\begin{equation}
\begin{aligned}
\Delta\hat{\mathbf X}_{0,t}^z
&=F_\theta^z(\mathbf Q_t^z,\mathbf G,\sigma_t),~~~
\mathbf v_t^{z,\mathrm{corr}}=\mathbf v_t^{z,\mathrm{base}}
-\frac{(1-\sigma_t)^2}{\sigma_t+\sigma_{\min}}
\Delta\hat{\mathbf X}_{0,t}^z,
\qquad z\in\{A,C\}.
\end{aligned}
\label{eq:maif_correction}
\end{equation}
Here $F_\theta^z$ comprises the shared MAIF trunk and the output head
for stream $z$, with $\theta$ denoting MAIF's trainable parameters.
The gate $(1-\sigma_t)^2$ suppresses corrections at high noise, where
clean estimates are less reliable, and fixed $\sigma_{\min}>0$
stabilizes the conversion. Residuals are assembled into each stream's
native layout, and corrected clean estimates are recovered as
$\hat{\mathbf X}_{0,t}^{z,\mathrm{corr}}
=\mathbf X_t^z-\sigma_t\mathbf v_t^{z,\mathrm{corr}}$.

\paragraph{Flow and physical supervision.}
We supervise each stream in normalized flow space and decoded physical
space. The fixed decoders recover end-effector positions and object-relative poses for ICT:
\begin{equation}
\begin{aligned}
\hat{\mathbf P}_t^{A,\mathrm{corr}}
&=\mathcal D_A\!\left(
\hat{\mathbf X}_{0,t}^{A,\mathrm{corr}}\right), ~~
(\hat{\mathbf P}_t^{C,\mathrm{corr}},
 \hat{\mathbf R}_t^{C,\mathrm{corr}})
=\mathcal D_C\!\left(
\hat{\mathbf X}_{0,t}^{C,\mathrm{corr}}\right).
\end{aligned}
\label{eq:maif_physical_decoding}
\end{equation}
Here $\mathbf P$ and $\mathbf R$ collect positions and rotations over
the horizon and arm slots. Gradients from the physical losses pass
through the fixed decoders to MAIF. Let $\mathcal I_A$ index valid
action coordinates $(b,h,j)$, with $j\in\{1,\ldots,D_a\}$ indexing
the native action vector; coordinates of unavailable arms and padding
are excluded. Let $\mathbf p_{bhr}^{A,*}$ denote the expert
end-effector position in the robot base frame. Using scalar velocity
components $v_{t,bhj}^{A,\mathrm{corr}}$ and $v_{bhj}^{A,*}$, the action
losses and adapter objective are
\begin{equation}
\begin{aligned}
\mathcal L_{\mathrm{flow}}^{A,\mathrm{corr}}
={}&\frac{1}{|\mathcal I_A|}
\sum_{(b,h,j)\in\mathcal I_A}
\left(
v_{t,bhj}^{A,\mathrm{corr}}-v_{bhj}^{A,*}
\right)^2,~~~
\mathcal L_{\mathrm{pos}}^{A,\mathrm{corr}}
={}\frac{1}{|\mathcal V|}
\sum_{(b,h,r)\in\mathcal V}
\rho_{\mathrm{SL1}}\!\left(
\frac{\hat{\mathbf p}_{t,bhr}^{A,\mathrm{corr}}
-\mathbf p_{bhr}^{A,*}}{s_{\mathrm{pos}}}
\right),
\\
\mathcal L_{\mathrm{MAIF}}
={}&\mathcal L_{\mathrm{flow}}^{A,\mathrm{corr}}
+\lambda_{\mathrm{pos}}\mathcal L_{\mathrm{pos}}^{A,\mathrm{corr}} +\lambda_{\mathrm{ICT}}\!\left(
\mathcal L_{\mathrm{flow}}^{C,\mathrm{corr}}
+\lambda_{\mathrm{pose}}\mathcal L_{\mathrm{pose}}^{C,\mathrm{corr}}
\right).
\end{aligned}
\label{eq:maif_objective}
\end{equation}
The action flow target $\mathbf v^{A,*}$ follows
Eq.~\eqref{eq:ict_flow_path}; its loss is an unweighted mean over
valid action coordinates. We set $s_{\mathrm{pos}}=0.01\,\mathrm m$.
The penalty $\rho_{\mathrm{SL1}}$ sums smooth-$L_1$ penalties over the
three normalized position residuals: a scalar residual $u$ contributes
$\frac12u^2$ for $|u|\leq1$ and $|u|-\frac12$ otherwise.

The two ICT terms reuse Eq.~\eqref{eq:ict_combined_loss} on corrected
predictions, with the same flow weighting, robust pose penalty, and
low-noise mask as Stage~1. The positive weights
$\lambda_{\mathrm{pos}}$ and $\lambda_{\mathrm{ICT}}$ scale action
position supervision and auxiliary ICT supervision, respectively.
The ICT losses train the shared MAIF trunk, which is also used by the
action branch at inference.

\paragraph{Inference.}
At inference, only the MAIF action branch is used. The native
scheduler updates actions with $\mathbf v_t^{A,\mathrm{corr}}$. Action and ICT tokens remain
mutually visible in the frozen expert throughout denoising, allowing
actions to use the generated ICT. Only the final action chunk is
executed.

\section{Experiments}
We conduct experiments in both simulation and real-world settings to evaluate the effectiveness of our proposed modules. In simulation, we evaluate our method on two widely used manipulation benchmarks, RoboTwin 2.0~\cite{robotwin2.0} and LIBERO~\cite{libero}. We further compare its performance and generalization on five real-world manipulation tasks against competitive VLA and WAM baselines. More training and experiment details can be found in the Appendix \ref{appendix}.

\subsection{RoboTwin 2.0 Simulation}
\paragraph{Evaluation protocol.}
We evaluate StarVLA~\cite{starvla}, SpatialVLA~\cite{spatialvla},
$\pi_{0.5}$~\cite{pi0.5}, Fast-WAM~\cite{fast-wam}, and
X-WAM~\cite{X-WAM} on all 50 RoboTwin 2.0 tasks.
These baselines span VLA and WAM families and differ in robot-data
pretraining and the use of explicit geometric inputs.
This selection assesses our framework's applicability across
backbone families and whether it provides additional gains for
geometry-aware policies.
All baselines and their ``+ Ours'' variants are trained on the same
50 clean demonstration episodes per task. ``+ Ours'' denotes ICT
adaptation followed by 5,000 MAIF training steps with the adapted
backbone frozen. Each method is evaluated for 100 episodes per task
in each setting (clean and randomized). We report task-averaged
success rates and their mean across the two settings (Avg.).
Training details are provided in the Appendix \ref{appendix}.

\paragraph{Results.}
Table~\ref{tab:robotwin_main} shows improvements in both clean and
randomized settings across all five baselines. Gains in average
success range from 2.24 to 5.94 percentage points, with a mean gain
of 3.59 points across baselines.
StarVLA, $\pi_{0.5}$, and Fast-WAM benefit more under randomization,
with gains of 4.62, 5.18, and 9.56 points, respectively, compared
with 1.92, 0.58, and 2.32 points in clean scenes. These results
suggest improved generalization from clean demonstrations to
randomized environments.
The geometry-aware SpatialVLA and X-WAM also improve by 2.24 and
3.60 points on average, respectively, suggesting that metric
interaction complements existing geometric conditioning.
X-WAM + Ours achieves the best among the evaluated
methods: 87.00\% in clean scenes, 77.02\% under randomization, and
82.01\% on average owing to its pretraining on RoboTwin 2.0. 

\begin{table}[t]
    \centering
    \setlength{\abovecaptionskip}{1pt}
    \setlength{\belowcaptionskip}{1pt}
    \caption{\textbf{Average success on 50 RoboTwin 2.0 tasks.} Best in \textbf{bold}; second-best \underline{underlined}.}
    \vspace{2pt}
    \label{tab:robotwin_main}
    \footnotesize
    \renewcommand{\arraystretch}{0.8}
    \resizebox{0.8\linewidth}{!}{
    \begin{tabular}{lccccc}
        \toprule
        Method & Robo. P.T. & Geo.Input & Clean (\%) & Randomized (\%) & Avg. (\%) \\
        \midrule
        StarVLA            & \xmark & \xmark & 72.54 & 14.64 & 43.59 \\
        SpatialVLA         & \cmark & \cmark & 3.14  & 1.62  & 2.38  \\
        $\pi_{0.5}$        & \cmark & \xmark & 70.68  & 46.0 & 58.34 \\
        Fast-WAM           & \xmark & \xmark & 77.76 & 1.92 & 39.84 \\
        X-WAM              & \cmark & \cmark & \underline{82.46} & \underline{74.36} & \underline{78.41} \\
        \midrule
        \textbf{StarVLA + Ours}     & \xmark & \cmark & 74.46 \mygreen{(+1.92)}  & 19.26 \mygreen{(+4.62)} & 46.86 \mygreen{(+3.27)}\\
        \textbf{SpatialVLA + Ours}  & \cmark & \cmark & 5.62  \mygreen{(+2.48)}  & 3.62  \mygreen{(+2.0)} & 4.62  \mygreen{(+2.24)}\\
        $\mathbf{\pi_{0.5}}$ + \textbf{Ours} & \cmark & \cmark & 71.26 \mygreen{(+0.58)} & 51.18 \mygreen{(+5.18)} & 61.22 \mygreen{(+2.88)}\\
        \textbf{Fast-WAM + Ours}    & \xmark & \cmark & 80.08 \mygreen{(+2.32)} & 11.48 \mygreen{(+9.56)} & 45.78 \mygreen{(+5.94)} \\
        \textbf{X-WAM + Ours}       & \cmark & \cmark & \textbf{87.0} \mygreen{(+4.54)} & \textbf{77.02} \mygreen{(+2.66)} & \textbf{82.01} \mygreen{(+3.6)}\\
        \bottomrule
    \end{tabular}}
    \vspace{-8pt}
\end{table}

\subsection{LIBERO Simulation}

\paragraph{Evaluation protocol.}
We evaluate our framework on LIBERO, using 50 evaluation episodes for each of the 10 tasks per suite.
The four suites are Spatial, Object, Goal, and Long.
We report success rates for each suite.
We also report their unweighted mean to summarize overall task performance.
Experiments use a WAM backbone (Fast-WAM) and a VLA backbone ($\pi_{0.5}$), following each model's native training and rollout protocol.
For this single-arm benchmark, one 9D ICT arm slot is active within the shared 18D interface, and the unused slot is masked.
Fast-WAM uses the video-action-visible training mask, allowing future-video queries to attend to action and ICT tokens. We disable bimanual attention during MAIF training on LIBERO.
\paragraph{Results.}
Table~\ref{tab:libero} shows higher average success with our framework on both backbones: from 96.75\% to 97.80\% for Fast-WAM and from 96.85\% to 97.40\% for $\pi_{0.5}$.
Both improve on Spatial and Long, while Fast-WAM gains 2.4 percentage points on Goal.
The gains vary across suites: Fast-WAM drops 0.6 percentage points on Object, and $\pi_{0.5}$ remains unchanged on Goal.

\begin{table}[t]
    \centering
    \setlength{\abovecaptionskip}{1pt}
    \setlength{\belowcaptionskip}{1pt}
    \caption{\textbf{Results} on LIBERO. Best in \textbf{bold}; second-best \underline{underlined}.}
    \vspace{2pt}
    \label{tab:libero}
    \footnotesize
    \renewcommand{\arraystretch}{0.8}
    \resizebox{0.8\linewidth}{!}{
    \begin{tabular}{lcccccc}
        \toprule
        Method & Robo. P.T. & Spatial (\%) & Object (\%) & Goal (\%) & Long (\%) & Avg. (\%) \\
        \midrule
        $\pi_{0.5}$        & \cmark & \underline{98.8} & 98.2             & \underline{98.0} & 92.4             & 96.85 \\
        Fast-WAM           & \xmark & 97.6             & \textbf{99.4}    & 96.0             & \underline{94.0} & 96.75 \\
        \midrule
        $\mathbf{\pi_{0.5}}$ + \textbf{Ours} & \cmark & \textbf{99.0} \mygreen{(+0.2)} & \underline{99.2} \mygreen{(+1.0)} & \underline{98.0} (=) & 93.4 \mygreen{(+1.0)}  & \underline{97.40} \mygreen{(+0.55)} \\
        \textbf{Fast-WAM + Ours}    & \xmark & \underline{98.8} \mygreen{(+1.2)} & 98.8 \myred{(-0.6)}& \textbf{98.4} \mygreen{(+2.4)} & \textbf{95.2} \mygreen{(+1.2)} & \textbf{97.80} \mygreen{(+1.05)} \\
        \bottomrule
    \end{tabular}}
    \vspace{-8pt}
\end{table}

\begin{figure}[!t]
  \centering
  \includegraphics[width=1.0\linewidth]{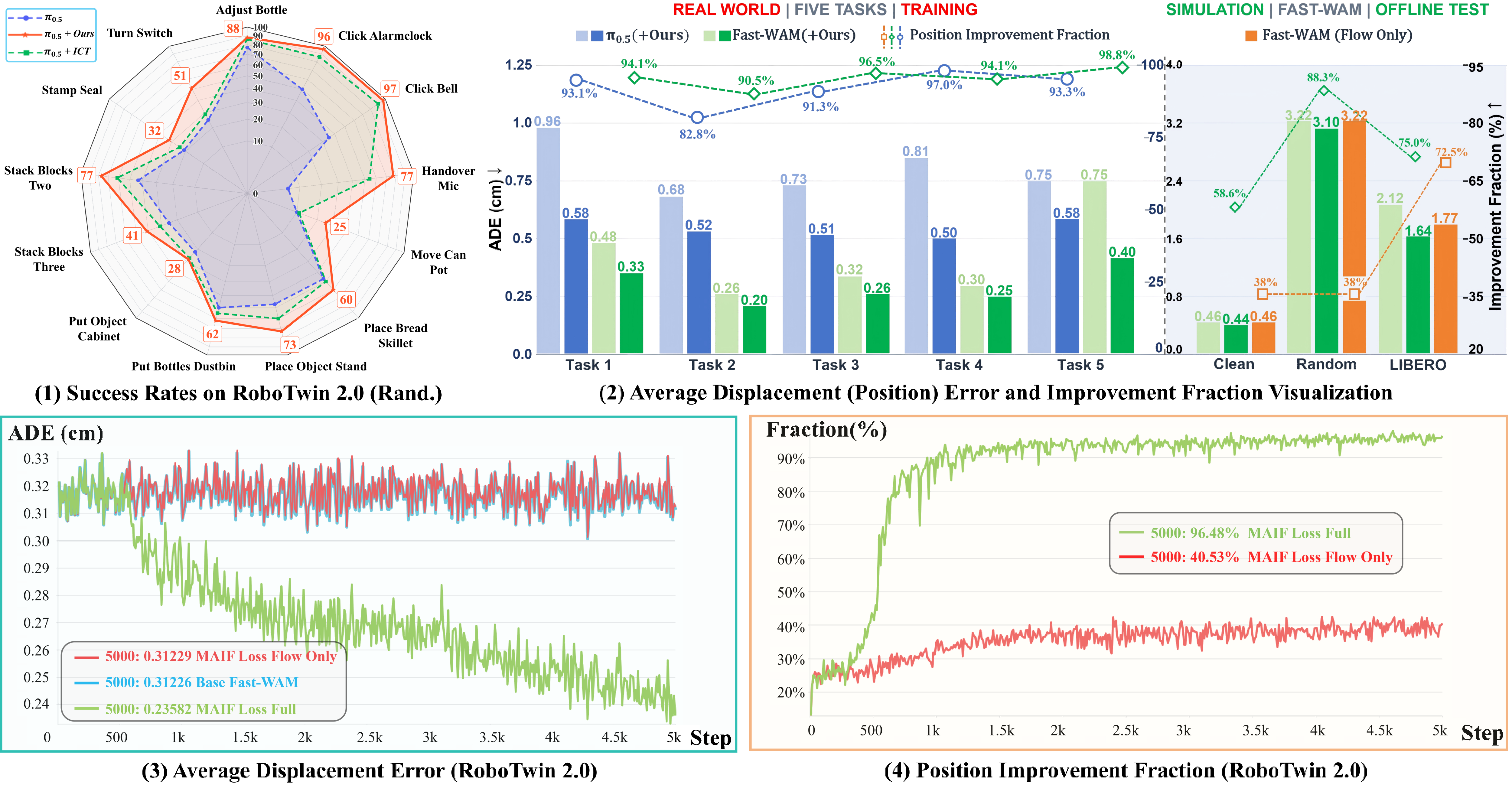}
  \caption{
\textbf{Experiment Details.}
(1) Random success rates on 13 tasks after clean-only training, showing that our method improves OOD generalization. (2) \textbf{Left:} Training results on real-world tasks and \textbf{Right:} Offline inference on simulation benchmarks. Bars show Average Displacement Error (ADE) and dashed lines indicate the fraction of samples with reduced ADE. Our method improves both backbones and consistently outperforms flow-only
version in training and testing metrics.
(3) and (4): Training logs. Flow-only supervision barely reduces the Average ADE of and improves the accuracy of the baseline's trajectory in only about 40\% of samples. While our method with the full loss incorporating waypoint and ICT supervision substantially reduces trajectory ADE and improves base-trajectory accuracy in over 95\% of samples.}
  \label{fig:Experiment_graph}
\end{figure}

\subsection{Ablation Study}

\begin{table}[t]
    \centering
    \setlength{\abovecaptionskip}{1pt}
    \setlength{\belowcaptionskip}{1pt}
    \caption{\textbf{Ablation Result} on RoboTwin 2.0. Best in \textbf{bold}; second-best \underline{underlined}.}
    \vspace{2pt}
    \label{tab:robotwin_ablation}
    \footnotesize
    \renewcommand{\arraystretch}{0.8}
    \resizebox{0.8\linewidth}{!}{
    \begin{tabular}{clccc}
        \toprule
        Type & Method & Clean (\%) & Randomized (\%) & Avg. (\%) \\
        \midrule
        \multirow{2}{*}{\textbf{VLA}}
        & $\pi_{0.5}$ + ICT
        & \underline{70.92} \mygreen{(+0.24)} & \underline{48.98} \mygreen{(+2.98)}& \underline{59.95} \mygreen{(+1.61)}\\
        & $\mathbf{\pi_{0.5}}$ + \textbf{Ours}
        & \textbf{71.26} \mygreen{(+0.58)} & \textbf{51.18} \mygreen{(+5.18)} & \textbf{61.22} \mygreen{(+2.88)} \\
        \midrule
        \multirow{8}{*}{\textbf{WAM}}
        & Fast-WAM + Flow Only
        & 78.08 & 2.64 & 40.36 \\
        & Fast-WAM 35k step
        & 77.92 & 3.34 & 40.63 \\
        & \textbf{\textit{Separate Coordinate Frames}}
        & 77.4 & 3.62 & 40.51 \\
        & $\mathbf{0.5} \times$ scale
        & 77.2 & 2.4 & 39.8 \\
        & $\mathbf{2.0} \times$ scale
        & 76.9 & 3.26 & 40.08 \\
        & \textbf{\textit{normalized}} scale
        & 77.68 & 2.8 & 40.24 \\
        & Fast-WAM + ICT
        & \underline{78.4} \mygreen{(+0.64)} & \underline{4.1} \mygreen{(+2.18)} & \underline{41.25} \mygreen{(+1.41)} \\
        & \textbf{Fast-WAM + Ours}
        & \textbf{80.08} \mygreen{(+2.32)} & \textbf{11.48} \mygreen{(+9.56)} & \textbf{45.78} \mygreen{(+5.94)}\\
        \bottomrule
    \end{tabular}}
    \vspace{-8pt}
\end{table}

We conduct ablation studies on RoboTwin 2.0 using $\pi_{0.5}$ and Fast-WAM to examine the contributions of interaction modeling, metric consistency, and supervision. Both baselines are trained exclusively on clean demonstrations for
30,000 steps with a global batch size of 256. We then train MAIF for 5,000 steps with the backbone frozen and a global batch size of 256. As Table~\ref{tab:robotwin_ablation} shows, adding ICT improves both backbones, while incorporating MAIF further increases the average success from $59.95\%$ to $61.22\%$ for $\pi_{0.5}$ and from $41.25\%$ to $45.78\%$ for Fast-WAM. The corresponding gains under randomization are $2.20$ and $7.38$ percentage points, respectively. 
Fig.~\ref{fig:Experiment_graph} (1) shows success rates on 13 tasks (Random) of $\pi_{0.5}$ and its variants with our method; our two-stage training consistently improves the success rate.
These improvements support the complementary roles of object-level interaction and scene-level action refinement to generalization.

To assess whether the gains merely arise from additional training, we continue training Fast-WAM on clean episodes for another $5000$ steps, matching the number of additional training steps. This baseline achieves $40.63\%$ average success, compared with $45.78\%$ for our full method, suggesting that longer training alone does not explain the improvement. We also evaluate \textbf{\textit{Separate Coordinate Frames}}, where point clouds
and end-effector trajectories are expressed in different coordinate
frames, alongside scene-wise point-cloud \textbf{\textit{normalization}} and scaling
by factors of $\mathbf{0.5}$ and $\mathbf{2.0}$. These variants achieve average success
rates of only $39.80\%$--$40.51\%$, highlighting the importance of
coordinate alignment and a consistent scale between
end-effector motion and scene geometry.

Finally, training MAIF with only action-velocity supervision yields $40.36\%$ average success below the full method's $45.78\%$. 
Fig.~\ref{fig:Experiment_graph}(3,4) compares the full loss objective with its
Flow Only variant. The further reduces the Average
Displacement Error (ADE) to $0.24\,\mathrm{cm}$,
whereas the latter remains close to the baseline at
$0.31\,\mathrm{cm}$. And the position improvement fraction reaches $96.48\%$ compared with $40.53\%$ for Flow Only.
The right part of Fig.~\ref{fig:Experiment_graph}(2) reports offline simulation
results. Across two settings of RoboTwin~2.0 and
LIBERO, the full method achieves lower ADE
and higher improvement fractions than Flow Only.
This suggests that velocity-space supervision alone provides insufficient guidance for accurate action refinement. Our full method uses physically grounded supervision for both
object- and scene-level interactions, encouraging more
geometrically consistent corrections to the denoising velocity
and enabling more precise manipulation.

\subsection{Real-World Experiments}

\begin{figure}[!t]
  \centering
  \includegraphics[width=1.0\linewidth]{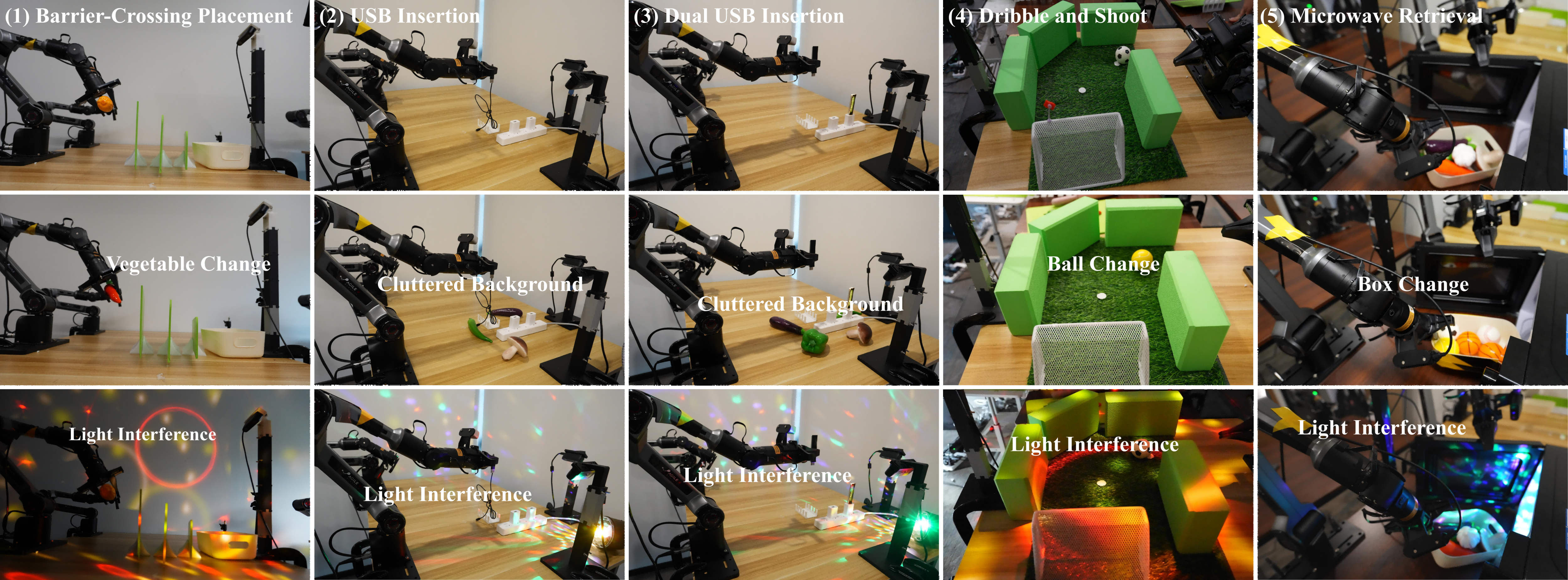}
  \caption{\textbf{Real-world task settings and out-of-distribution (OOD) perturbations.} Each column corresponds to one of the five manipulation tasks. The top row shows standard task configurations; the middle row presents OOD.1 scenarios with object variations or background clutter; and the bottom row shows OOD.2 scenarios under lighting perturbations.}
  \label{fig:Experiment}
\end{figure}

We evaluate two baselines Fast-WAM~\cite{fast-wam} and $\pi_{0.5}$~\cite{pi0.5} on a dual-arm Piper-H platform equipped with a fixed-view head camera and wrist cameras, all using Orbbec Gemini 335 RGB-D sensors. The baseline policies use observations from all cameras, whereas
MAIF constructs scene geometry from the fixed-view camera's depth
observations. Further details on the real-world setup, data collection, and training
are provided in the Appendix \ref{appendix}.

We evaluate four single-arm tasks and one bimanual task as Fig.~\ref{fig:Experiment} shows. \textbf{Task 1: Barrier-Crossing Corn Placement}, lifting an ear of corn over barriers of varying heights and placing it into a bowl; \textbf{Task 2: USB Insertion}, aligning and inserting a USB connector into its port; \textbf{Task 3: Dual-USB Insertion}, sequentially inserting two USB connectors into their corresponding ports; \textbf{Task 4: Obstacle-Course Soccer}, dribbling a ball around obstacles before shooting it into a goal; and \textbf{Task 5: Bimanual Microwave Retrieval}, using one arm to open the door and retrieve a box while the other pushes the door aside. Together, these tasks require height perception, fine-grained spatial alignment, obstacle-aware planning, and depth perception. All tasks are evaluated 100 times.

\paragraph{(1) Standard Task Evaluation.} 
The left half of Fig.~\ref{fig:Experiment_graph}(2) reports training metrics on five tasks. Our method substantially reduces ADE for both
$\pi_{0.5}$ and Fast-WAM across all tasks.
The position improvement fraction exceeds $90\%$ in every
task and backbone except $\pi_{0.5}$ + Ours on
Task 2 ($82.8\%$).
These results indicate more accurate end-effector trajectory
predictions, supporting more precise real-world manipulation.
Table~\ref{tab:real_world} reports the success rates under the standard setting. Adding our two-level metric interaction framework consistently improves both policy backbones. Fast-WAM improves from $40.8\%$ to $46.8\%$, corresponding to a gain of $6.0$ percentage points, while $\pi_{0.5}$ improves from $69.4\%$ to $77.0\%$, a gain of $7.6$ points. The improvements are particularly pronounced on the more geometry-sensitive tasks. On Dual-USB Insertion, our method improves Fast-WAM and $\pi_{0.5}$ by $6$ and $12$ points, respectively. On Bimanual Microwave Retrieval, the gains reach $13$ and $15$ points. These tasks require precise relative-pose estimation, constrained-space reasoning, and coordinated action refinement, directly reflecting the complementary roles of object-level ICT and scene-level MAIF. Although both $\pi_{0.5}$ and $\pi_{0.5}$ with our framework achieve $100\%$ success on Task 1, our method clears the medium-height barrier with a lower trajectory (a more rational pattern). Detailed experiment visualizations are provided in the Appendix \ref{appendix}.

\paragraph{(2) Out-of-Distribution (OOD) Study.} We further evaluate robustness under two types of unseen environmental perturbations: task-specific scene or background changes and lighting changes shown in Fig.~\ref{fig:Experiment}. These variations modify visual appearance and scene context while preserving the underlying task objective, allowing us to assess whether the policy relies on appearance correlations or transferable metric interaction cues.
As shown in Table~\ref{tab:realworld_ood}, our method improves every task--condition pair for both backbones. The average OOD success rate of Fast-WAM increases from $15.1\%$ to $22.6\%$, while that of $\pi_{0.5}$ increases from $50.0\%$ to $57.4\%$, yielding gains of $7.5$ and $7.4$ percentage points. The largest improvements occur on Task 3: Fast-WAM improves from $9\%$ to $24\%$ under scene perturbation and from $0\%$ to $15\%$ under lighting changes, while $\pi_{0.5}$ improves from $36\%$ to $51\%$ and from $24\%$ to $37\%$. These results indicate that explicitly grounding end-effector--object relations and action trajectories in a shared metric space improves not only nominal task success but also transfer and generalization under out-of-distribution environmental changes.

\begin{table}[t]
    \centering
    \setlength{\abovecaptionskip}{1pt}
    \setlength{\belowcaptionskip}{1pt}
    \caption{\textbf{Real-World Success Rate} on Five Tasks. Best in \textbf{bold}; second-best \underline{underlined}.}
    \vspace{2pt}
    \label{tab:real_world}
    \footnotesize
    \renewcommand{\arraystretch}{0.92}
    \resizebox{0.8\linewidth}{!}{
    \begin{tabular}{lcccccc}
        \toprule
        Method & Task 1(\%) & Task 2(\%) & Task 3(\%) & Task 4(\%) & Task 5(\%) &  Average(\%)\\
        \midrule
        Fast-WAM             & 75             & 19             & 30             & 23             & 57             & 40.8 \\
        $\mathbf{\pi_{0.5}}$          & \textbf{100}   & \underline{74} & \underline{47} & \underline{61} & 65             & \underline{69.4} \\
        \textbf{Fast-WAM + Ours}      & \underline{81} & 21             & 36             & 26             & \underline{70} & 46.8 \mygreen{(+6.0)} \\        
        $\mathbf{\pi_{0.5}}$ + \textbf{Ours}   & \textbf{100}   & \textbf{81}    & \textbf{59}    & \textbf{65}    & \textbf{80}    & \textbf{77.0} \mygreen{(+7.6)}\\
        \bottomrule
    \end{tabular}}
    \vspace{-8pt}
\end{table}

\begin{table*}[t]
\centering
\setlength{\abovecaptionskip}{1pt}
\setlength{\belowcaptionskip}{1pt}
\caption{\textbf{Real-world OOD success rates}. Best in \textbf{bold}; second-best \underline{underlined}.}
\vspace{2pt}
\label{tab:realworld_ood}
\footnotesize
\renewcommand{\arraystretch}{0.92}
\setlength{\tabcolsep}{3pt}

\resizebox{\textwidth}{!}{
\begin{tabular}{lcc|cc|cc|cc|cc|c}
    \toprule
    Method
    & \multicolumn{2}{c|}{Task 1 (\%)}
    & \multicolumn{2}{c|}{Task 2 (\%)}
    & \multicolumn{2}{c|}{Task 3 (\%)}
    & \multicolumn{2}{c|}{Task 4 (\%)}
    & \multicolumn{2}{c|}{Task 5 (\%)}
    & Summary (\%)
    \\
    
    \cmidrule(lr){2-3}
    \cmidrule(lr){4-5}
    \cmidrule(lr){6-7}
    \cmidrule(lr){8-9}
    \cmidrule(lr){10-11}
    \cmidrule(lr){12-12}
    
    & OOD.1 & OOD.2 
    & OOD.1 & OOD.2 
    & OOD.1 & OOD.2 
    & OOD.1 & OOD.2 
    & OOD.1 & OOD.2
    & Average
    \\
    
    \midrule
    
    Fast-WAM
    & 43 & 39
    & 13 & 12
    & 9  & 0
    & 11 & 7
    & 15 & 2
    & 15.1
    \\
    
    $\mathbf{\pi_{0.5}}$
    & \underline{92} & \underline{90}
    & \underline{58} & \underline{49}
    & \underline{36} & \underline{24}
    & \underline{34} & \underline{25}
    & \underline{41} & \underline{51}
    & \underline{50.0} 
    \\

    \textbf{Fast-WAM + Ours}
    & 44 & 46
    & 19 & 19
    & 24 & 15
    & 12 & 15
    & 22 & 10
    & 22.6 \mygreen{(+7.5)}
    \\
    
    $\mathbf{\pi_{0.5}}$ + \textbf{Ours}
    & \textbf{95} & \textbf{97}
    & \textbf{62} & \textbf{53}
    & \textbf{51} & \textbf{37}
    & \textbf{40} & \textbf{32}
    & \textbf{48} & \textbf{59}
    & \textbf{57.4} \mygreen{(+7.4)}
    \\

    \bottomrule
\end{tabular}
}

\vspace{-8pt}
\end{table*}

\section{Conclusion}
We presented a metric interaction framework in Cartesian space
for precise robotic manipulation. ICTs provide object-relative end-effector
pose supervision and MAIF uses metric scene geometry to refine actions.
Our framework improves pose accuracy, task success and OOD generalization
across VLA and WAM baselines with few additional parameters and
training steps.

\newpage

\subsection*{AI use statement}

Generative AI tools were used to assist with writing and refining code,
preparing visualization scripts and figure layouts, formatting LaTeX tables,
and polishing the language and presentation of the manuscript. All
AI-assisted code and manuscript modifications were manually reviewed and
verified by the authors. The reported numerical results and result figures
were derived exclusively from experiments conducted by the authors; no
experimental data, numerical results, or result figures were synthetically
generated or fabricated using generative AI. The authors take full
responsibility for the accuracy and integrity of the final paper.

\bibliography{iclr2027_conference}
\bibliographystyle{iclr2027_conference}

\newpage

\appendix
\section{Appendix}
\label{appendix}
\subsection{Sonata Feature Fusion}

\begin{figure}[h]
    \centering
    \includegraphics[width=\linewidth]{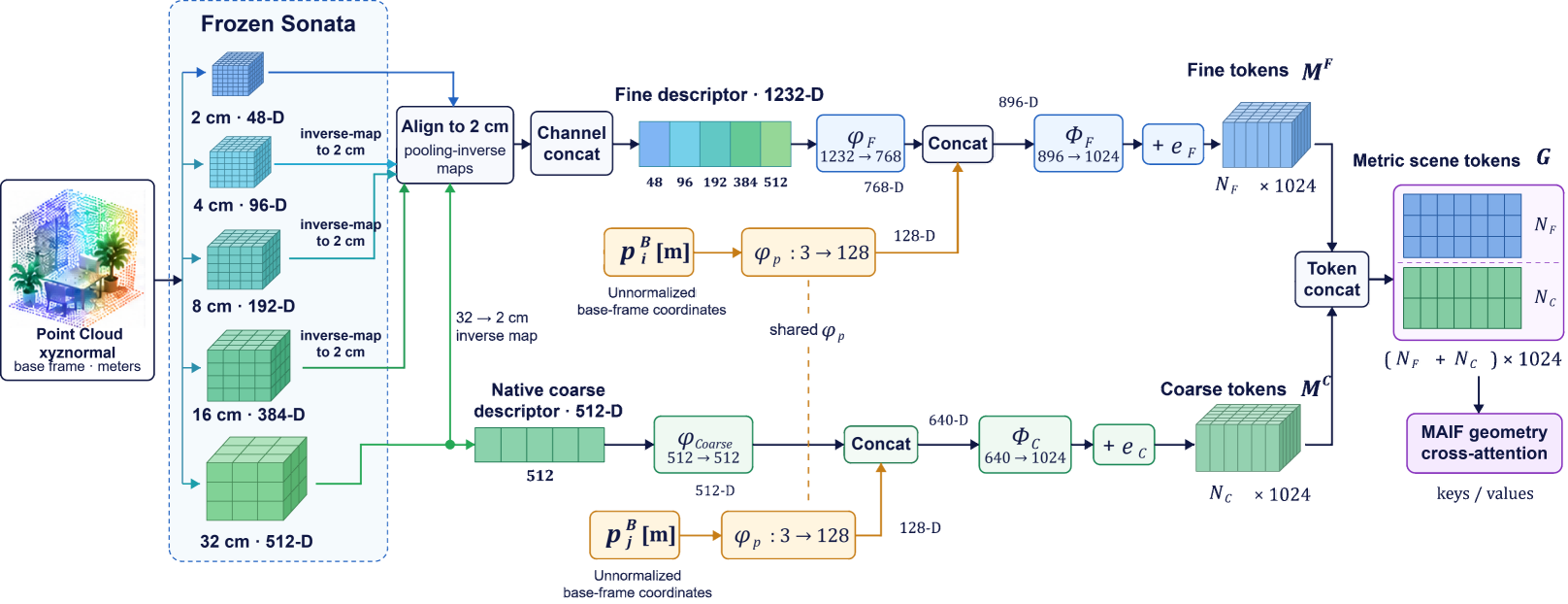}
    \caption{
    \textbf{Construction of multi-scale metric scene tokens.}
    A frozen Sonata encoder processes point coordinates and surface
    normals without RGB attributes.
    Features from all five levels are aligned to the finest
    $2\,\mathrm{cm}$ voxels and channel-concatenated into
    1232-dimensional fine descriptors, while native
    $32\,\mathrm{cm}$ features form a parallel coarse stream.
    Each stream fuses its descriptors with embeddings of the
    corresponding unnormalized robot-base-frame coordinates
    in meters.
    The resulting 1024-dimensional fine and coarse tokens are
    concatenated to form $\mathbf G$, which supplies the keys
    and values for MAIF geometry cross-attention.
    }
    \label{fig:multiscale_metric_memory}
\end{figure}

\paragraph{Multi-scale Sonata feature aggregation.}
We construct multi-scale metric scene tokens from a frozen
Sonata encoder~\cite{sonata}.
Across all experiments, the encoder receives 3D point coordinates
and surface normals, without RGB attributes.
The encoder produces a five-level sparse hierarchy with voxel
sizes of $(2,4,8,16,32)\,\mathrm{cm}$ and corresponding feature
dimensions of $(48,96,192,384,512)$.
Let $\mathbf F^s$ denote the feature matrix at voxel size
$s\,\mathrm{cm}$, with $\mathbf f_i^s$ denoting its $i$-th row.

To construct the fine stream ($F$), we follow Sonata's
feature-upcasting scheme and use its pooling-inverse maps to
lift every coarser feature to the finest $2\,\mathrm{cm}$ voxels.
The aligned features are then concatenated along the channel
dimension:
\begin{equation}
\begin{aligned}
\bar{\mathbf f}_i^{F}
={}&
\bigl[
\mathbf f_i^{2};
\bigl(\mathcal U_{4\rightarrow2}(\mathbf F^{4})\bigr)_i;
\bigl(\mathcal U_{8\rightarrow2}(\mathbf F^{8})\bigr)_i;
\\
&\quad
\bigl(\mathcal U_{16\rightarrow2}(\mathbf F^{16})\bigr)_i;
\bigl(\mathcal U_{32\rightarrow2}(\mathbf F^{32})\bigr)_i
\bigr]
\in \mathbb R^{1232}.
\end{aligned}
\label{eq:sonata_multiscale_aggregation}
\end{equation}
Here, $\mathcal U_{s\rightarrow2}$ denotes lifting through the
composed pooling-inverse maps.
The native $32\,\mathrm{cm}$ features are also retained as a
parallel 512-dimensional coarse stream ($C$).

Inspired by the feature--position concatenation used in
SpatialLM~\cite{spatiallm}, we separately project each stream's
geometric descriptor and its corresponding base-frame coordinate,
then concatenate and fuse the two embeddings.
Let $\mathbf p_i^B$ and $\mathbf p_j^B$ denote the coordinates
associated with the fine and coarse descriptors, respectively,
where $B$ denotes the robot base frame.
For this coordinate branch, we retain the original coordinates
in meters without normalization:
\begin{equation}
\begin{aligned}
\mathbf m_i^{F}
&=
\Phi_F\!\left(
\left[
\phi_F(\bar{\mathbf f}_i^{F});
\phi_p(\mathbf p_i^B)
\right]
\right)
+\mathbf e_F,
\\
\mathbf m_j^{C}
&=
\Phi_C\!\left(
\left[
\phi_{\mathrm{coarse}}(\mathbf f_j^{32});
\phi_p(\mathbf p_j^B)
\right]
\right)
+\mathbf e_C.
\end{aligned}
\label{eq:metric_feature_fusion}
\end{equation}
The descriptor projections
$\phi_F:\mathbb R^{1232}\rightarrow\mathbb R^{768}$ and
$\phi_{\mathrm{coarse}}:\mathbb R^{512}\rightarrow\mathbb R^{512}$
embed the fine and coarse descriptors, respectively.
The shared coordinate projection
$\phi_p:\mathbb R^3\rightarrow\mathbb R^{128}$
produces a 128-dimensional coordinate embedding for each stream.
The fusion projections
$\Phi_F:\mathbb R^{896}\rightarrow\mathbb R^{1024}$ and
$\Phi_C:\mathbb R^{640}\rightarrow\mathbb R^{1024}$
map the concatenated embeddings to a common token dimension
$D=1024$.
The learned embeddings
$\mathbf e_F,\mathbf e_C\in\mathbb R^{1024}$
identify the fine and coarse streams.

Finally, we concatenate the two streams along the token dimension:
\begin{equation}
\mathbf G
=
\mathbf M^B
=
\operatorname{Concat}_{\mathrm{token}}
\left(\mathbf M^F,\mathbf M^C\right)
\in\mathbb R^{(N_F+N_C)\times1024}.
\label{eq:metric_point_memory}
\end{equation}
Here, $\mathbf M^F$ and $\mathbf M^C$ stack the fine and coarse
tokens, respectively, and $N_F$ and $N_C$ denote their token counts.
The resulting memory provides the keys and values for MAIF
geometry cross-attention.
As illustrated in Fig~\ref{fig:multiscale_metric_memory},
this construction combines local detail with coarse scene context
and explicitly incorporates metric position through unnormalized
base-frame coordinates.

\subsection{RoboTwin 2.0 Training Details}

\paragraph{Simulation backbone preparation.}
We evaluate MAIF with five representative policy backbones:
Fast-WAM~\cite{fast-wam}, $\pi_{0.5}$~\cite{pi0.5},
X-WAM~\cite{X-WAM}, StarVLA~\cite{starvla}, and
SpatialVLA~\cite{spatialvla}.
All policies follow the same RoboTwin 2.0 evaluation protocol,
including environment action semantics, language prompts,
success criteria, and rollout rules.
We retain each policy's native temporal parameterization:
Fast-WAM, $\pi_{0.5}$, X-WAM, and StarVLA use an action horizon
of 32, whereas SpatialVLA predicts and executes a four-step
action chunk.

For Fast-WAM and $\pi_{0.5}$, our local reproductions closely
matched the public RoboTwin leaderboard results within expected
rollout variation.
We therefore report their public leaderboard success rates
as baseline values.
All other baselines and all MAIF variants are evaluated locally.

The released X-WAM RoboTwin checkpoint uses a different
training-data protocol from our clean-only setting.
We therefore initialize from its released pretrained checkpoint,
whose world-model backbone originates from Wan2.2-TI2V-5B,
and fine-tune it for 30,000 steps on 2,500 clean demonstrations
(50 tasks with 50 demonstrations per task).
Backbone fine-tuning uses eight GPUs, a per-GPU batch size
of 32, and no gradient accumulation.

We use the released PI-style configuration,
\texttt{VLAct\_Qwen3PI\_Robotwin\_Finetune}, denoted
\emph{StarVLA-PI} for StarVLA.
It combines the 100K-step VLAct-pretrained Qwen3-VL-4B action
backbone with a Qwen3-0.6B \texttt{QwenPI\_v4} flow-matching
action expert.
We follow the released initialization recipe and fine-tune
for 50,000 steps on the clean-only
\texttt{robotwin\_wrap\_32} dataset.
We evaluate this exact variant locally because the public
headline result uses a different action head.

For SpatialVLA, we initialize from the released
\texttt{spatialvla-4b-224-pt} checkpoint and perform non-LoRA
fine-tuning for 30,000 steps on the same 2,500 clean demonstrations.
The policy consumes one head-camera RGB image and predicts
four 14-D absolute bimanual joint/gripper actions.
Continuous action dimensions use training-set
$q_{0.01}$--$q_{0.99}$ normalization, while grippers remain binary.
The token embeddings and internal ZoeDepth estimator are frozen;
the vision tower and remaining policy parameters are trainable.
Backbone fine-tuning uses RGB without oracle simulator depth,
eight GPUs, a global batch size of 256, BF16, and ZeRO-1.
The learning rate is $2\times10^{-5}$ with zero weight decay,
150 warmup steps, and linear decay.
We evaluate the resulting 30K checkpoint locally.

\paragraph{MAIF training in simulation.}
For every backbone, we freeze the complete task-fine-tuned
policy and the Sonata encoder and optimize only newly
initialized MAIF parameters.
All adapters are trained on the same 2,500 clean demonstrations
for 5,000 optimizer steps.
Oracle simulator depth observations are back-projected into
the robot base frame and sampled into 8,192 points.
Point coordinates and action waypoints share the same metric
frame without per-scene centering, scale normalization,
or test-time alignment. For all RoboTwin experiments, the policy except SpatialVLA retains its native three-view RGB
input, whereas MAIF uses only the static head-camera oracle depth to construct metric point cloud in the robot base frame. While to align with the official monocular input configuration, SpatialVLA uses only the head camera.

The common MAIF architecture uses a hidden dimension of 1,024,
16 attention heads, three geometry cross-attention layers,
ten temporal interaction layers, and two bimanual interaction
layers.
Training uses AdamW with a peak learning rate of $10^{-4}$,
weight decay $10^{-4}$, gradient clipping at $1.0$,
BF16 mixed precision, and random seed 42.
Unless otherwise specified, we use a 500-step linear warmup
followed by cosine decay.
Backbone-dependent batch sizes and parameter counts are
summarized in Table~\ref{tab:robotwin_maif_impl}.

In our MAIF pipeline, the frozen X-WAM forward pass remains
memory-intensive because it processes additional conditioning
inputs, including depth and robot state.
Together with MAIF training, this limits the micro-batch to
one window per GPU.
We therefore use four-step gradient accumulation across
eight GPUs, giving an effective global batch of 32 windows.

For SpatialVLA, MAIF refines the decoded four-step action chunk
using frozen action-token features and a zero-initialized
residual in normalized action space.
Its external geometry stream uses oracle simulator depth.


\begin{table*}[t]
    \centering
    \caption{\textbf{Backbone preparation and MAIF training on RoboTwin 2.0.}
    All backbones are fine-tuned on clean demonstrations.
    Batch sizes count distinct scene--action windows;
    parameter counts refer to one deployed MAIF adapter.}
    \label{tab:robotwin_maif_impl}

    \footnotesize
    \renewcommand{\arraystretch}{1.15}
    \setlength{\tabcolsep}{4pt}

    \begin{tabularx}{\textwidth}{
        @{}
        >{\raggedright\arraybackslash}p{0.14\textwidth}
        >{\raggedright\arraybackslash}X
        >{\centering\arraybackslash}p{0.13\textwidth}
        >{\centering\arraybackslash}p{0.15\textwidth}
        @{}
    }
        \multicolumn{4}{@{}l@{}}{
            \textbf{(a) Backbone preparation}
        } \\[3pt]
        \toprule
        \textbf{Backbone}
            & \textbf{RoboTwin policy initialization}
            & \shortstack{\textbf{Fine-tuning}\\\textbf{steps}}
            & \shortstack{\textbf{Baseline SR}\\\textbf{source}} \\
        \midrule

        Fast-WAM
            & Released Fast-WAM initialization
            & 30K
            & Leaderboard \\
        \addlinespace[3pt]

        $\pi_{0.5}$
            & Released \texttt{pi05\_base} initialization
            & 30K
            & Leaderboard \\
        \addlinespace[3pt]

        X-WAM
            & Released X-WAM pretrained checkpoint with
              a Wan2.2-TI2V-5B backbone
            & 30K
            & Local \\
        \addlinespace[3pt]

        StarVLA-PI
            & VLAct/Qwen3-VL-4B action backbone with
              a Qwen3-0.6B \texttt{QwenPI\_v4} expert
            & 50K
            & Local \\
        \addlinespace[3pt]

        SpatialVLA
            & Released \texttt{spatialvla-4b-224-pt};
              PaLiGemma2 backbone pretrained on OXE and RH20T
            & 30K
            & Local \\

        \bottomrule
    \end{tabularx}

    \par\vspace{8pt}

    \begin{tabularx}{\textwidth}{
        @{}
        >{\raggedright\arraybackslash}p{0.14\textwidth}
        >{\centering\arraybackslash}p{0.13\textwidth}
        >{\raggedright\arraybackslash}X
        >{\centering\arraybackslash}p{0.17\textwidth}
        >{\centering\arraybackslash}p{0.13\textwidth}
        @{}
    }
        \multicolumn{5}{@{}l@{}}{
            \textbf{(b) MAIF training}
        } \\[3pt]
        \toprule
        \textbf{Backbone}
            & \shortstack{\textbf{Windows}\\\textbf{/ GPU}}
            & \shortstack[l]{\textbf{Resampling /}\\\textbf{accumulation}}
            & \shortstack{\textbf{Global windows}\\\textbf{/ step}}
            & \shortstack{\textbf{MAIF}\\\textbf{params}} \\
        \midrule

        Fast-WAM
            & 32
            & None
            & 256
            & 198.2M \\
        \addlinespace[3pt]

        $\pi_{0.5}$
            & 32
            & None
            & 256
            & 198.2M \\
        \addlinespace[3pt]

        X-WAM
            & 1
            & 4-step gradient accumulation
            & 32
            & 200.6M \\
        \addlinespace[3pt]

        StarVLA-PI
            & 8
            & 4 flow-time/noise pairs per window
            & 64
            & 198.2M \\
        \addlinespace[3pt]

        SpatialVLA
            & 1
            & None
            & 8
            & 202.1M \\

        \bottomrule
    \end{tabularx}

    \par\vspace{4pt}

    \begin{minipage}{\textwidth}
        \scriptsize
        \raggedright

        \textit{Evaluation.}
        ``Leaderboard'' indicates that the baseline success rate
        is taken from the public benchmark after confirming
        consistency with our local evaluation.
        All MAIF results are evaluated locally.

        \par\smallskip
        \textit{Batch interpretation.}
        All configurations use eight GPUs.
        Windows/GPU counts distinct windows loaded before
        resampling or accumulation; global windows/step
        includes gradient accumulation.
        StarVLA-PI uses four independently sampled
        flow-time/noise pairs per window, yielding
        256 flow-supervision pairs per optimizer step.

        \par\smallskip
        \textit{Parallel ablations.}
        SpatialVLA trains eight independent adapters jointly
        on the same input batch and cached backbone features.
        These branches are evaluated individually and are
        not combined at deployment.
    \end{minipage}
\end{table*}

\subsection{LIBERO Training Details}
\label{sec:libero_ict_training}

\paragraph{Training data.}
We jointly train on the no-op-filtered versions of LIBERO-Spatial,
LIBERO-Object, LIBERO-Goal, and LIBERO-10. The combined training set contains
1,712 demonstrations and 277,713 temporal windows, comprising 53,229,
67,309, 52,895, and 104,280 windows from the four suites, respectively.
No additional validation split is constructed.

Each sample contains 33 consecutive observations and the following 32 robot
actions at 20 Hz. We retain video frames at temporal indices
$\{0,4,8,\ldots,32\}$, resulting in nine frames per sample. The external-view
and wrist-view RGB images are independently resized from $512\times512$ to
$224\times224$ and concatenated horizontally, producing an input of shape
$[B,3,9,224,448]$. Images remain in RGB order throughout preprocessing.
The action target has seven dimensions: a 6-DoF relative end-effector command
and one absolute gripper command. The proprioceptive state has eight dimensions,
consisting of the 6-DoF end-effector pose and two gripper joint positions.
Actions and states are normalized using fixed training-set min--max statistics.

We use cached language embeddings instead of running the text encoder during
training. Each instruction is represented by a sequence of 128 tokens with a
feature dimension of 4096. The proprioceptive state at the first observation
is projected into an additional context token and appended to the language
sequence.

\paragraph{Optimization.}
We train the complete VideoDiT--ActionDiT MoT and the proprioceptive encoder,
while freezing the VAE and omitting the online text encoder. Training uses
eight GPUs with a per-GPU batch size of 16, gradient accumulation of 1, and
a global batch size of 128. We use bfloat16 precision and DeepSpeed ZeRO-2
without parameter or optimizer offloading. The optimizer is AdamW with
$\beta_1=0.9$, $\beta_2=0.95$, learning rate $10^{-4}$, weight decay $10^{-2}$,
and gradient-norm clipping at 1.0. The learning rate uses linear warm-up
followed by cosine annealing with a minimum learning rate of $10^{-6}$.
Gradient checkpointing is disabled.

Starting from optimization step 0, we train for 30,000 optimizer steps,
corresponding to approximately 3.84 million sampled windows or 13.8 passes
over the combined LIBERO training set. We use the final checkpoint at
step 30,000 for evaluation.

\paragraph{Policy inference.}
At test time, the policy jointly denoises a 32-step action chunk and its ICT
trajectory for 10 flow-matching steps, conditioned only on the current
two-view observation, proprioception, and language instruction. The
training-only future-video-to-action/ICT edges are disabled, and no future
video is generated. The environment executes the first 10 predicted actions
before replanning. We disable action ensembling and binarize the predicted
gripper command.

\paragraph{LIBERO MAIF training.}
For both Fast-WAM and $\pi_{0.5}$, Stage~2 MAIF training starts
from the corresponding policy checkpoint obtained after
Stage~1 ICT training.
The complete ICT-trained policy and the Sonata point encoder
remain frozen, and only the newly initialized MAIF adapter,
containing approximately 200M trainable parameters, is optimized.

The original policy observations are preserved: both the fixed
\textit{agentview} and eye-in-hand RGB streams are provided
to the policy, while only the calibrated \textit{agentview}
depth observation is used to construct metric geometry.
Depth is back-projected into the MuJoCo world frame and
sampled into an 8,192-point cloud.
Point coordinates and predicted end-effector waypoints are
represented in the same metric coordinate system without
per-scene centering, scale normalization, or test-time alignment.
We retain the native action horizon of each backbone,
using 32 steps for Fast-WAM and 10 steps for $\pi_{0.5}$.

Both implementations follow the same MAIF optimization protocol:
5,000 optimizer steps on eight GPUs with a per-GPU batch size
of 32 (global batch size 256) and no gradient accumulation.
We use AdamW with a peak learning rate of $10^{-4}$,
weight decay $10^{-4}$, gradient clipping at 1.0,
and a 500-step linear warmup followed by cosine decay
to $10^{-6}$.
Training uses BF16 mixed precision and random seed 42.
For Fast-WAM, frozen video-VAE latents and text embeddings
are cached offline to reduce training overhead.

\subsection{RoboTwin 2.0 Detailed Results}
\begin{table*}[t]
\centering
\tiny
\caption{Per-task success rates on RoboTwin under clean and randomized evaluation settings. }
\label{tab:robotwin_detail_1}
\setlength{\tabcolsep}{2pt}
\resizebox{\textwidth}{!}{
\begin{tabular}{lcc|cc|cc|cc|cc|cc}
\toprule
Task  &  \multicolumn{2}{c}{Fast-WAM}  &  \multicolumn{2}{c}{Fast-WAM + ICT}  &  \multicolumn{2}{c}{Fast-WAM + Ours}  &  \multicolumn{2}{c}{\makecell{$\pi_{0.5}$}}  &  \multicolumn{2}{c}{$\pi_{0.5}$ + ICT}  &  \multicolumn{2}{c}{$\pi_{0.5}$ + Ours}  \\
\cmidrule(lr){2-3}\cmidrule(lr){4-5}\cmidrule(lr){6-7}\cmidrule(lr){8-9}\cmidrule(lr){10-11}\cmidrule(lr){12-13}
  &  Clean  &  Rand.  &  Clean  &  Rand.  &  Clean  &  Rand.  &  Clean  &  Rand.  &  Clean  &  Rand.  &  Clean  &  Rand.  \\
\midrule
Adjust Bottle               & 99   & 0  & 100  & 0   & 100  & 6   & 99   & 77  & 98  & 86  & 99   & 88    \\
Beat Block Hammer           & 82   & 0  & 97   & 0   & 86   & 2   & 93   & 23  & 65  & 31  & 58   & 19    \\
Blocks Ranking RGB          & 81   & 0  & 93   & 0   & 89   & 0   & 72   & 44  & 76  & 61  & 88   & 48    \\
Blocks Ranking Size         & 52   & 0  & 63   & 0   & 63   & 0   & 45   & 21  & 46  & 33  & 44   & 18    \\
Click Alarmclock            & 99   & 60 & 100  & 39  & 99   & 52  & 65   & 50  & 100 & 86  & 100  & 96    \\
Click Bell                  & 95   & 7  & 100  & 30  & 97   & 76  & 31   & 35  & 100 & 90  & 100  & 97    \\
Dump Bin Bigbin             & 94   & 0  & 93   & 2   & 89   & 0   & 92   & 84  & 88  & 82  & 90   & 80    \\
Grab Roller                 & 100  & 2  & 100  & 0   & 100  & 7   & 99   & 97  & 99  & 89  & 100  & 92    \\
Handover Block              & 66   & 0  & 54   & 0   & 25   & 0   & 30   & 12  & 54  & 10  & 10   & 14    \\
Handover Mic                & 98   & 0  & 99   & 0   & 95   & 0   & 98   & 6   & 99  & 54  & 92   & 77    \\
Hanging Mug                 & 14   & 0  & 35   & 0   & 20   & 0   & 17   & 14  & 13  & 13  & 14   & 3     \\
Lift Pot                    & 91   & 0  & 91   & 0   & 87   & 0   & 98   & 35  & 67  & 19  & 86   & 47    \\
Move Can Pot                & 96   & 0  & 73   & 0   & 97   & 2   & 60   & 10  & 53  & 11  & 77   & 25    \\
Move Pillbottle Pad         & 92   & 0  & 92   & 0   & 94   & 0   & 46   & 32  & 60  & 28  & 63   & 69    \\
Move Playingcard Away       & 94   & 0  & 96   & 0   & 92   & 22  & 84   & 65  & 84  & 72  & 82   & 70    \\
Move Stapler Pad            & 33   & 0  & 24   & 1   & 25   & 0   & 20   & 18  & 32  & 9   & 18   & 7     \\
Open Laptop                 & 84   & 0  & 96   & 0   & 94   & 20  & 93   & 68  & 98  & 60  & 91   & 62    \\
Open Microwave              & 17   & 0  & 15   & 0   & 31   & 80  & 86   & 59  & 38  & 17  & 67   & 23    \\
Pick Diverse Bottles        & 72   & 0  & 61   & 0   & 86   & 0   & 66   & 42  & 72  & 42  & 69   & 50    \\
Pick Dual Bottles           & 87   & 1  & 75   & 3   & 100  & 0   & 72   & 57  & 74  & 50  & 86   & 67    \\
Place A2B Left              & 69   & 0  & 77   & 0   & 72   & 1   & 71   & 49  & 70  & 36  & 61   & 31    \\
Place A2B Right             & 60   & 1  & 71   & 1   & 82   & 1   & 68   & 39  & 68  & 37  & 67   & 37    \\
Place Bread Basket          & 87   & 0  & 89   & 1   & 83   & 0   & 63   & 60  & 76  & 68  & 68   & 50    \\
Place Bread Skillet         & 85   & 0  & 89   & 0   & 92   & 0   & 76   & 47  & 68  & 50  & 72   & 60    \\
Place Burger Fries          & 97   & 0  & 100  & 1   & 97   & 0   & 82   & 86  & 88  & 84  & 92   & 86    \\
Place Can Basket            & 77   & 0  & 46   & 0   & 91   & 0   & 56   & 10  & 51  & 25  & 41   & 20    \\
Place Cans Plasticbox       & 98   & 0  & 93   & 2   & 100  & 0   & 31   & 73  & 66  & 70  & 89   & 48    \\
Place Container Plate       & 100  & 0  & 99   & 6   & 98   & 0   & 92   & 72  & 92  & 79  & 97   & 73    \\
Place Dual Shoes            & 67   & 0  & 72   & 0   & 70   & 0   & 70   & 45  & 63  & 56  & 66   & 34    \\
Place Empty Cup             & 90   & 0  & 95   & 1   & 92   & 0   & 96   & 80  & 81  & 64  & 90   & 69    \\
Place Fan                   & 59   & 0  & 65   & 0   & 57   & 0   & 66   & 26  & 56  & 37  & 39   & 31    \\
Place Mouse Pad             & 55   & 0  & 63   & 1   & 78   & 0   & 34   & 31  & 38  & 19  & 47   & 19    \\
Place Object Basket         & 82   & 0  & 71   & 0   & 66   & 1   & 70   & 29  & 73  & 37  & 47   & 22    \\
Place Object Scale          & 83   & 0  & 77   & 0   & 76   & 0   & 78   & 38  & 66  & 40  & 64   & 39    \\
Place Object Stand          & 94   & 0  & 94   & 4   & 95   & 0   & 85   & 47  & 83  & 60  & 94   & 73    \\
Place Phone Stand           & 85   & 0  & 70   & 0   & 82   & 0   & 65   & 37  & 47  & 28  & 51   & 38    \\
Place Shoe                  & 87   & 0  & 93   & 7   & 95   & 0   & 82   & 50  & 87  & 72  & 83   & 63    \\
Press Stapler               & 90   & 3  & 96   & 24  & 93   & 91  & 79   & 63  & 92  & 52  & 86   & 61    \\
Put Bottles Dustbin         & 55   & 0  & 38   & 0   & 77   & 0   & 78   & 50  & 75  & 55  & 72   & 62    \\
Put Object Cabinet          & 42   & 0  & 48   & 0   & 82   & 0   & 39   & 22  & 44  & 27  & 58   & 28    \\
Rotate QRcode               & 81   & 0  & 79   & 0   & 67   & 1   & 90   & 21  & 76  & 21  & 44   & 23    \\
Scan Object                 & 68   & 0  & 83   & 1   & 69   & 0   & 49   & 40  & 56  & 39  & 25   & 20    \\
Shake Bottle                & 100  & 10 & 100  & 35  & 100  & 79  & 100  & 100 & 100 & 92  & 99   & 99    \\
Shake Bottle Horizontally   & 100  & 9  & 100  & 29  & 100  & 75  & 100  & 100 & 100 & 94  & 99   & 99    \\
Stack Blocks Three          & 89   & 0  & 77   & 0   & 62   & 0   & 77   & 25  & 73  & 31  & 74   & 41    \\
Stack Blocks Two            & 99   & 0  & 95   & 0   & 98   & 0   & 91   & 43  & 80  & 61  & 93   & 77    \\
Stack Bowls Three           & 80   & 0  & 68   & 0   & 62   & 0   & 81   & 52  & 58  & 47  & 75   & 51    \\
Stack Bowls Two             & 94   & 0  & 92   & 3   & 88   & 0   & 94   & 70  & 90  & 72  & 96   & 70    \\
Stamp Seal                  & 30   & 0  & 74   & 0   & 62   & 0   & 53   & 21  & 55  & 24  & 80   & 32   \\
Turn Switch                 & 39   & 3  & 48   & 14  & 50   & 58  & 52   & 25  & 58  & 29  & 60   & 51    \\
\midrule
\textbf{Average}             & 77.76 & 1.92 & \underline{78.40} & 4.1 & \textbf{80.08} & 11.48 & 70.68 & 46.0  & 70.92  & \underline{48.98}  & 71.26 & \textbf{51.18}   \\
\bottomrule
\end{tabular}
}
\end{table*}

\begin{table*}[t]
\centering
\tiny
\caption{Per-task success rates on RoboTwin under clean and randomized evaluation settings.}
\label{tab:robotwin_detail_2}
\setlength{\tabcolsep}{2pt}
\resizebox{\textwidth}{!}{
\begin{tabular}{lcc|cc|cc|cc|cc|cc}
\toprule
Task
& \multicolumn{2}{c}{StarVLA}
& \multicolumn{2}{c}{StarVLA + Ours}
& \multicolumn{2}{c}{SpatialVLA}
& \multicolumn{2}{c}{SpatialVLA + Ours}
& \multicolumn{2}{c}{X-WAM}
& \multicolumn{2}{c}{X-WAM + Ours}
\\
\cmidrule(lr){2-3}
\cmidrule(lr){4-5}
\cmidrule(lr){6-7}
\cmidrule(lr){8-9}
\cmidrule(lr){10-11}
\cmidrule(lr){12-13}
& Clean & Rand.
& Clean & Rand.
& Clean & Rand.
& Clean & Rand.
& Clean & Rand.
& Clean & Rand.
\\
\midrule
Adjust Bottle               & 91  & 0   & 99  & 19  & 1   & 0   & 1   & 1   & 99  & 98  & 100 & 100 \\
Beat Block Hammer           & 75  & 19  & 79  & 22  & 0   & 0   & 0   & 0   & 98  & 29  & 97  & 42  \\
Blocks Ranking RGB          & 58  & 0   & 82  & 9   & 0   & 0   & 0   & 0   & 89  & 90  & 94  & 85  \\
Blocks Ranking Size         & 51  & 0   & 45  & 0   & 0   & 0   & 0   & 0   & 71  & 42  & 55  & 51  \\
Click Alarmclock            & 43  & 19  & 33  & 78  & 11  & 0   & 77  & 0   & 99  & 91  & 100 & 99  \\
Click Bell                  & 96  & 10  & 95  & 94  & 9   & 0   & 32  & 11  & 100 & 97  & 100 & 98  \\
Dump Bin Bigbin             & 81  & 71  & 96  & 56  & 0   & 0   & 12  & 0   & 95  & 90  & 97  & 91  \\
Grab Roller                 & 97  & 25  & 96  & 58  & 11  & 0   & 0   & 0   & 100 & 75  & 100 & 74  \\
Handover Block              & 89  & 0   & 87  & 0   & 0   & 0   & 0   & 0   & 69  & 74  & 91  & 71  \\
Handover Mic                & 86  & 11  & 99  & 3   & 0   & 0   & 0   & 0   & 78  & 61  & 87  & 85  \\
Hanging Mug                 & 20  & 0   & 16  & 0   & 0   & 0   & 0   & 0   & 20  & 40  & 44  & 39  \\
Lift Pot                    & 95  & 0   & 96  & 0   & 0   & 0   & 0   & 0   & 99  & 88  & 99  & 89  \\
Move Can Pot                & 35  & 0   & 80  & 0   & 2   & 0   & 1   & 0   & 61  & 79  & 86  & 90  \\
Move Pillbottle Pad         & 78  & 0   & 90  & 0   & 0   & 0   & 0   & 0   & 90  & 91  & 98  & 98  \\
Move Playingcard Away       & 90  & 0   & 99  & 0   & 2   & 0   & 4   & 0   & 89  & 66  & 100 & 85  \\
Move Stapler Pad            & 40  & 0   & 37  & 0   & 0   & 0   & 0   & 0   & 38  & 35  & 60  & 33  \\
Open Laptop                 & 88  & 0   & 97  & 21  & 4   & 5   & 1   & 0   & 92  & 85  & 90  & 80  \\
Open Microwave              & 55  & 19  & 9   & 94  & 15  & 31  & 13   & 37  & 68  & 48  & 64  & 52  \\
Pick Diverse Bottles        & 57  & 61  & 66  & 60  & 0   & 0   & 0   & 0   & 91  & 80  & 94  & 81  \\
Pick Dual Bottles           & 90  & 45  & 65  & 67  & 0   & 0   & 0   & 0   & 100 & 99  & 100 & 100 \\
Place A2B Left              & 57  & 0   & 62  & 0   & 1   & 0   & 1   & 0   & 78  & 30  & 81  & 43  \\
Place A2B Right             & 49  & 4   & 39  & 0   & 0   & 0   & 0   & 0   & 59  & 77  & 82  & 52  \\
Place Bread Basket          & 77  & 24  & 70  & 30  & 0   & 0   & 0   & 0   & 92  & 82  & 90  & 90  \\
Place Bread Skillet         & 79  & 5   & 91  & 14  & 0   & 0   & 0   & 0   & 88  & 88  & 95  & 93  \\
Place Burger Fries          & 91  & 0   & 95  & 7   & 0   & 0   & 0   & 0   & 97  & 87  & 95  & 89  \\
Place Can Basket            & 45  & 0   & 77  & 0   & 0   & 0   & 0   & 0   & 50  & 71  & 61  & 59  \\
Place Cans Plasticbox       & 99  & 9   & 98  & 0   & 0   & 0   & 0   & 0   & 100 & 89  & 99  & 96  \\
Place Container Plate       & 99  & 44  & 94  & 51  & 0   & 0   & 25  & 0   & 98  & 93  & 99  & 97  \\
Place Dual Shoes            & 86  & 37  & 75  & 0   & 0   & 0   & 0   & 0   & 70  & 91  & 82  & 90  \\
Place Empty Cup             & 92  & 7   & 96  & 6   & 0   & 0   & 0   & 0   & 98  & 90  & 97  & 97  \\
Place Fan                   & 71  & 0   & 81  & 0   & 0   & 0   & 0   & 0   & 79  & 69  & 87  & 82  \\
Place Mouse Pad             & 45  & 0   & 57  & 0   & 0   & 0   & 0   & 0   & 65  & 65  & 89  & 62  \\
Place Object Basket         & 78  & 0   & 65  & 0   & 0   & 0   & 0   & 0   & 85  & 55  & 80  & 31  \\
Place Object Scale          & 74  & 12  & 56  & 11  & 0   & 0   & 0   & 0   & 95  & 56  & 88  & 61  \\
Place Object Stand          & 88  & 46  & 90  & 55  & 0   & 0   & 0   & 0   & 94  & 86  & 95  & 91  \\
Place Phone Stand           & 66  & 0   & 51  & 0   & 0   & 0   & 0   & 0   & 87  & 40  & 83  & 39  \\
Place Shoe                  & 90  & 40  & 80  & 24  & 0   & 0   & 0   & 0   & 87  & 75  & 97  & 93  \\
Press Stapler               & 85  & 48  & 87  & 45  & 50  & 24  & 46  & 39  & 93  & 77  & 95  & 93  \\
Put Bottles Dustbin         & 72  & 33  & 79  & 26  & 0   & 0   & 0   & 0   & 86  & 91  & 87  & 88  \\
Put Object Cabinet          & 70  & 0   & 43  & 0   & 0   & 0   & 0   & 0   & 74  & 57  & 75  & 62  \\
Rotate QRcode               & 81  & 0   & 84  & 0   & 0   & 0   & 0   & 0   & 80  & 73  & 81  & 52  \\
Scan Object                 & 80  & 0   & 96  & 0   & 0   & 0   & 0   & 0   & 71  & 58  & 75  & 63  \\
Shake Bottle                & 99  & 42  & 100 & 0   & 27  & 5   & 33  & 21  & 100 & 100 & 100 & 100 \\
Shake Bottle Horizontally   & 99  & 50  & 99  & 0   & 21  & 9  & 29  & 42  & 100 & 100 & 100 & 99  \\
Stack Blocks Three          & 45  & 0   & 64  & 0   & 0   & 0   & 0   & 0   & 71  & 83  & 77  & 81  \\
Stack Blocks Two            & 86  & 0   & 97  & 0   & 0   & 0   & 0   & 23  & 92  & 77  & 74  & 65  \\
Stack Bowls Three           & 55  & 11  & 68  & 0   & 0   & 0   & 0   & 0   & 70  & 85  & 84  & 87  \\
Stack Bowls Two             & 83  & 40  & 89  & 37  & 0   & 0   & 0   & 0   & 89  & 90  & 92  & 94  \\
Stamp Seal                  & 46  & 0   & 51  & 0   & 0   & 0   & 0   & 0   & 86  & 67  & 93  & 88  \\
Turn Switch                 & 25  & 0   & 23  & 76  & 3   & 7   & 6   & 7   & 43  & 58  & 61  & 71  \\
\midrule
\textbf{Average}
& 72.54\% & 14.64\%
& 74.46\% & 19.26\%
& 3.14\%  & 1.62\%
& 5.62\%  & 3.62\% 
& \underline{82.46\%} & \underline{74.36\%}
& \textbf{87.0\%} & \textbf{77.02\%}
\\
\bottomrule
\end{tabular}
}
\end{table*}
Tables~\ref{tab:robotwin_detail_1} and~\ref{tab:robotwin_detail_2} report the results of baseline methods and their variants augmented with our framework on RoboTwin 2.0. Each method is evaluated over 100 trials per task under the \textit{clean} and \textit{randomized} (Rand.) setting. RoboTwin 2.0 comprises 50 single-arm and bimanual manipulation tasks. For each task, we strictly follow the maximum rollout step specified
in the official open-source code. A trial is counted as a failure if
the task is not completed within this limit.


\subsection{LIBERO Detailed Results}
\label{app:libero_detailed_results}

The following tables report task-level results for Fast-WAM + Ours and
$\pi_{0.5}$ + Ours across all 40 tasks in LIBERO-Spatial, LIBERO-Object,
LIBERO-Goal, and LIBERO-10.
Each model is evaluated over 50 episodes per task, totaling 2,000 episodes
per model.
Fast-WAM + Ours succeeds in 1,956 episodes, achieving an overall success
rate of 97.8\%.
$\pi_{0.5}$ + Ours succeeds in 1,948 episodes, achieving an overall success
rate of 97.4\%.


\begingroup
\small
\setlength{\tabcolsep}{4pt}
\renewcommand{\arraystretch}{1.08}

\begin{longtable}{
    @{}
    >{\raggedright\arraybackslash}p{0.075\linewidth}
    >{\raggedright\arraybackslash}p{0.585\linewidth}
    >{\raggedleft\arraybackslash}p{0.075\linewidth}
    >{\raggedleft\arraybackslash}p{0.070\linewidth}
    >{\raggedleft\arraybackslash}p{0.095\linewidth}
    @{}
}
\caption{Task-level success rates on the four LIBERO benchmark suites.
Each task is evaluated over 50 episodes.  Task-level LIBERO results of Fast-WAM + Ours.}
\label{tab:libero_task_results} \\

\toprule
\textbf{Task} &
\textbf{Language instruction} &
\textbf{Succ.} &
\textbf{Trials} &
\textbf{Rate} \\
\midrule
\endfirsthead

\multicolumn{5}{c}{
    \tablename~\thetable{}: Task-level LIBERO results of Fast-WAM + Ours  (continued).
} \\
\toprule
\textbf{Task} &
\textbf{Language instruction} &
\textbf{Succ.} &
\textbf{Trials} &
\textbf{Rate} \\
\midrule
\endhead

\midrule
\multicolumn{5}{r}{\footnotesize Continued on the next page} \\
\endfoot

\bottomrule
\endlastfoot

\multicolumn{5}{@{}l}{\textbf{LIBERO-Spatial}} \\*
\addlinespace[0.15em]

S-0 &
Pick up the black bowl between the plate and the ramekin and place it on the plate. &
49 & 50 & 98.0\% \\

S-1 &
Pick up the black bowl next to the ramekin and place it on the plate. &
50 & 50 & 100.0\% \\

S-2 &
Pick up the black bowl from table center and place it on the plate. &
50 & 50 & 100.0\% \\

S-3 &
Pick up the black bowl on the cookie box and place it on the plate. &
50 & 50 & 100.0\% \\

S-4 &
Pick up the black bowl in the top drawer of the wooden cabinet and place it on the plate. &
49 & 50 & 98.0\% \\

S-5 &
Pick up the black bowl on the ramekin and place it on the plate. &
48 & 50 & 96.0\% \\

S-6 &
Pick up the black bowl next to the cookie box and place it on the plate. &
50 & 50 & 100.0\% \\

S-7 &
Pick up the black bowl on the stove and place it on the plate. &
50 & 50 & 100.0\% \\

S-8 &
Pick up the black bowl next to the plate and place it on the plate. &
49 & 50 & 98.0\% \\

S-9 &
Pick up the black bowl on the wooden cabinet and place it on the plate. &
49 & 50 & 98.0\% \\

\addlinespace[0.35em]
\multicolumn{2}{r}{\textit{LIBERO-Spatial subtotal}} &
\textbf{494} & \textbf{500} & \textbf{98.8\%} \\

\midrule

\multicolumn{5}{@{}l}{\textbf{LIBERO-Object}} \\*
\addlinespace[0.15em]

O-0 &
Pick up the alphabet soup and place it in the basket. &
50 & 50 & 100.0\% \\

O-1 &
Pick up the cream cheese and place it in the basket. &
50 & 50 & 100.0\% \\

O-2 &
Pick up the salad dressing and place it in the basket. &
50 & 50 & 100.0\% \\

O-3 &
Pick up the BBQ sauce and place it in the basket. &
49 & 50 & 98.0\% \\

O-4 &
Pick up the ketchup and place it in the basket. &
47 & 50 & 94.0\% \\

O-5 &
Pick up the tomato sauce and place it in the basket. &
50 & 50 & 100.0\% \\

O-6 &
Pick up the butter and place it in the basket. &
49 & 50 & 98.0\% \\

O-7 &
Pick up the milk and place it in the basket. &
49 & 50 & 98.0\% \\

O-8 &
Pick up the chocolate pudding and place it in the basket. &
50 & 50 & 100.0\% \\

O-9 &
Pick up the orange juice and place it in the basket. &
50 & 50 & 100.0\% \\

\addlinespace[0.35em]
\multicolumn{2}{r}{\textit{LIBERO-Object subtotal}} &
\textbf{494} & \textbf{500} & \textbf{98.8\%} \\

\midrule

\multicolumn{5}{@{}l}{\textbf{LIBERO-Goal}} \\*
\addlinespace[0.15em]

G-0 &
Open the middle drawer of the cabinet. &
48 & 50 & 96.0\% \\

G-1 &
Put the bowl on the stove. &
50 & 50 & 100.0\% \\

G-2 &
Put the wine bottle on top of the cabinet. &
50 & 50 & 100.0\% \\

G-3 &
Open the top drawer and put the bowl inside. &
49 & 50 & 98.0\% \\

G-4 &
Put the bowl on top of the cabinet. &
50 & 50 & 100.0\% \\

G-5 &
Push the plate to the front of the stove. &
50 & 50 & 100.0\% \\

G-6 &
Put the cream cheese in the bowl. &
50 & 50 & 100.0\% \\

G-7 &
Turn on the stove. &
50 & 50 & 100.0\% \\

G-8 &
Put the bowl on the plate. &
50 & 50 & 100.0\% \\

G-9 &
Put the wine bottle on the rack. &
45 & 50 & 90.0\% \\

\addlinespace[0.35em]
\multicolumn{2}{r}{\textit{LIBERO-Goal subtotal}} &
\textbf{492} & \textbf{500} & \textbf{98.4\%} \\

\midrule

\multicolumn{5}{@{}l}{\textbf{LIBERO-10}} \\*
\addlinespace[0.15em]

L10-0 &
Put both the alphabet soup and the tomato sauce in the basket. &
47 & 50 & 94.0\% \\

L10-1 &
Put both the cream cheese box and the butter in the basket. &
50 & 50 & 100.0\% \\

L10-2 &
Turn on the stove and put the moka pot on it. &
49 & 50 & 98.0\% \\

L10-3 &
Put the black bowl in the bottom drawer of the cabinet and close it. &
47 & 50 & 94.0\% \\

L10-4 &
Put the white mug on the left plate and put the yellow and white mug on the right plate. &
46 & 50 & 92.0\% \\

L10-5 &
Pick up the book and place it in the back compartment of the caddy. &
49 & 50 & 98.0\% \\

L10-6 &
Put the white mug on the plate and put the chocolate pudding to the right of the plate. &
43 & 50 & 86.0\% \\

L10-7 &
Put both the alphabet soup and the cream cheese box in the basket. &
48 & 50 & 96.0\% \\

L10-8 &
Put both moka pots on the stove. &
49 & 50 & 98.0\% \\

L10-9 &
Put the yellow and white mug in the microwave and close it. &
48 & 50 & 96.0\% \\

\addlinespace[0.35em]
\multicolumn{2}{r}{\textit{LIBERO-10 subtotal}} &
\textbf{476} & \textbf{500} & \textbf{95.2\%} \\

\midrule
\addlinespace[0.15em]
\multicolumn{2}{r}{\textbf{Overall}} &
\textbf{1,956} &
\textbf{2,000} &
\textbf{97.8\%} \\

\end{longtable}
\endgroup


\begingroup
\small
\setlength{\tabcolsep}{4pt}
\renewcommand{\arraystretch}{1.08}

\begin{longtable}{
    @{}
    >{\raggedright\arraybackslash}p{0.075\linewidth}
    >{\raggedright\arraybackslash}p{0.585\linewidth}
    >{\raggedleft\arraybackslash}p{0.075\linewidth}
    >{\raggedleft\arraybackslash}p{0.070\linewidth}
    >{\raggedleft\arraybackslash}p{0.095\linewidth}
    @{}
}
\caption{Task-level LIBERO results of $\pi_{0.5}$+Ours}
\label{tab:pi05_ict_libero_task_results} \\

\toprule
\textbf{Task} &
\textbf{Language instruction} &
\textbf{Succ.} &
\textbf{Trials} &
\textbf{Rate} \\
\midrule
\endfirsthead

\multicolumn{5}{c}{
    \tablename~\thetable{}: Task-level $\pi_{0.5}$+Ours results (continued)
} \\
\toprule
\textbf{Task} &
\textbf{Language instruction} &
\textbf{Succ.} &
\textbf{Trials} &
\textbf{Rate} \\
\midrule
\endhead

\midrule
\multicolumn{5}{r}{\footnotesize Continued on the next page} \\
\endfoot

\bottomrule
\endlastfoot

\multicolumn{5}{@{}l}{\textbf{LIBERO-Spatial}} \\*
\addlinespace[0.15em]

S-0 &
Pick up the black bowl between the plate and the ramekin and place it on the plate. &
50 & 50 & 100.0\% \\

S-1 &
Pick up the black bowl next to the ramekin and place it on the plate. &
50 & 50 & 100.0\% \\

S-2 &
Pick up the black bowl from table center and place it on the plate. &
50 & 50 & 100.0\% \\

S-3 &
Pick up the black bowl on the cookie box and place it on the plate. &
49 & 50 & 98.0\% \\

S-4 &
Pick up the black bowl in the top drawer of the wooden cabinet and place it on the plate. &
50 & 50 & 100.0\% \\

S-5 &
Pick up the black bowl on the ramekin and place it on the plate. &
49 & 50 & 98.0\% \\

S-6 &
Pick up the black bowl next to the cookie box and place it on the plate. &
50 & 50 & 100.0\% \\

S-7 &
Pick up the black bowl on the stove and place it on the plate. &
49 & 50 & 98.0\% \\

S-8 &
Pick up the black bowl next to the plate and place it on the plate. &
49 & 50 & 98.0\% \\

S-9 &
Pick up the black bowl on the wooden cabinet and place it on the plate. &
49 & 50 & 98.0\% \\

\addlinespace[0.35em]
\multicolumn{2}{r}{\textit{LIBERO-Spatial subtotal}} &
\textbf{495} & \textbf{500} & \textbf{99.0\%} \\

\midrule

\multicolumn{5}{@{}l}{\textbf{LIBERO-Object}} \\*
\addlinespace[0.15em]

O-0 &
Pick up the alphabet soup and place it in the basket. &
50 & 50 & 100.0\% \\

O-1 &
Pick up the cream cheese and place it in the basket. &
50 & 50 & 100.0\% \\

O-2 &
Pick up the salad dressing and place it in the basket. &
50 & 50 & 100.0\% \\

O-3 &
Pick up the BBQ sauce and place it in the basket. &
49 & 50 & 98.0\% \\

O-4 &
Pick up the ketchup and place it in the basket. &
48 & 50 & 96.0\% \\

O-5 &
Pick up the tomato sauce and place it in the basket. &
49 & 50 & 98.0\% \\

O-6 &
Pick up the butter and place it in the basket. &
50 & 50 & 100.0\% \\

O-7 &
Pick up the milk and place it in the basket. &
50 & 50 & 100.0\% \\

O-8 &
Pick up the chocolate pudding and place it in the basket. &
50 & 50 & 100.0\% \\

O-9 &
Pick up the orange juice and place it in the basket. &
50 & 50 & 100.0\% \\

\addlinespace[0.35em]
\multicolumn{2}{r}{\textit{LIBERO-Object subtotal}} &
\textbf{496} & \textbf{500} & \textbf{99.2\%} \\

\midrule

\multicolumn{5}{@{}l}{\textbf{LIBERO-Goal}} \\*
\addlinespace[0.15em]

G-0 &
Open the middle drawer of the cabinet. &
50 & 50 & 100.0\% \\

G-1 &
Put the bowl on the stove. &
50 & 50 & 100.0\% \\

G-2 &
Put the wine bottle on top of the cabinet. &
48 & 50 & 96.0\% \\

G-3 &
Open the top drawer and put the bowl inside. &
48 & 50 & 96.0\% \\

G-4 &
Put the bowl on top of the cabinet. &
50 & 50 & 100.0\% \\

G-5 &
Push the plate to the front of the stove. &
49 & 50 & 98.0\% \\

G-6 &
Put the cream cheese in the bowl. &
47 & 50 & 94.0\% \\

G-7 &
Turn on the stove. &
50 & 50 & 100.0\% \\

G-8 &
Put the bowl on the plate. &
50 & 50 & 100.0\% \\

G-9 &
Put the wine bottle on the rack. &
47 & 50 & 94.0\% \\

\addlinespace[0.35em]
\multicolumn{2}{r}{\textit{LIBERO-Goal subtotal}} &
\textbf{490} & \textbf{500} & \textbf{98.0\%} \\

\midrule

\multicolumn{5}{@{}l}{\textbf{LIBERO-10}} \\*
\addlinespace[0.15em]

L10-0 &
Put both the alphabet soup and the tomato sauce in the basket. &
50 & 50 & 100.0\% \\

L10-1 &
Put both the cream cheese box and the butter in the basket. &
50 & 50 & 100.0\% \\

L10-2 &
Turn on the stove and put the moka pot on it. &
47 & 50 & 94.0\% \\

L10-3 &
Put the black bowl in the bottom drawer of the cabinet and close it. &
49 & 50 & 98.0\% \\

L10-4 &
Put the white mug on the left plate and put the yellow and white mug on the right plate. &
50 & 50 & 100.0\% \\

L10-5 &
Pick up the book and place it in the back compartment of the caddy. &
49 & 50 & 98.0\% \\

L10-6 &
Put the white mug on the plate and put the chocolate pudding to the right of the plate. &
46 & 50 & 92.0\% \\

L10-7 &
Put both the alphabet soup and the cream cheese box in the basket. &
45 & 50 & 90.0\% \\

L10-8 &
Put both moka pots on the stove. &
38 & 50 & 76.0\% \\

L10-9 &
Put the yellow and white mug in the microwave and close it. &
44 & 50 & 88.0\% \\

\addlinespace[0.35em]
\multicolumn{2}{r}{\textit{LIBERO-10 subtotal}} &
\textbf{467} & \textbf{500} & \textbf{93.4\%} \\

\midrule
\addlinespace[0.15em]

\multicolumn{2}{r}{\textbf{Overall}} &
\textbf{1,948} &
\textbf{2,000} &
\textbf{97.40\%} \\

\end{longtable}
\endgroup

\subsection{Real-World Experiment Settings}
\label{app:real_world_experiment_settings}

\begin{figure}[!h]
  \centering
  \includegraphics[width=1.0\linewidth]{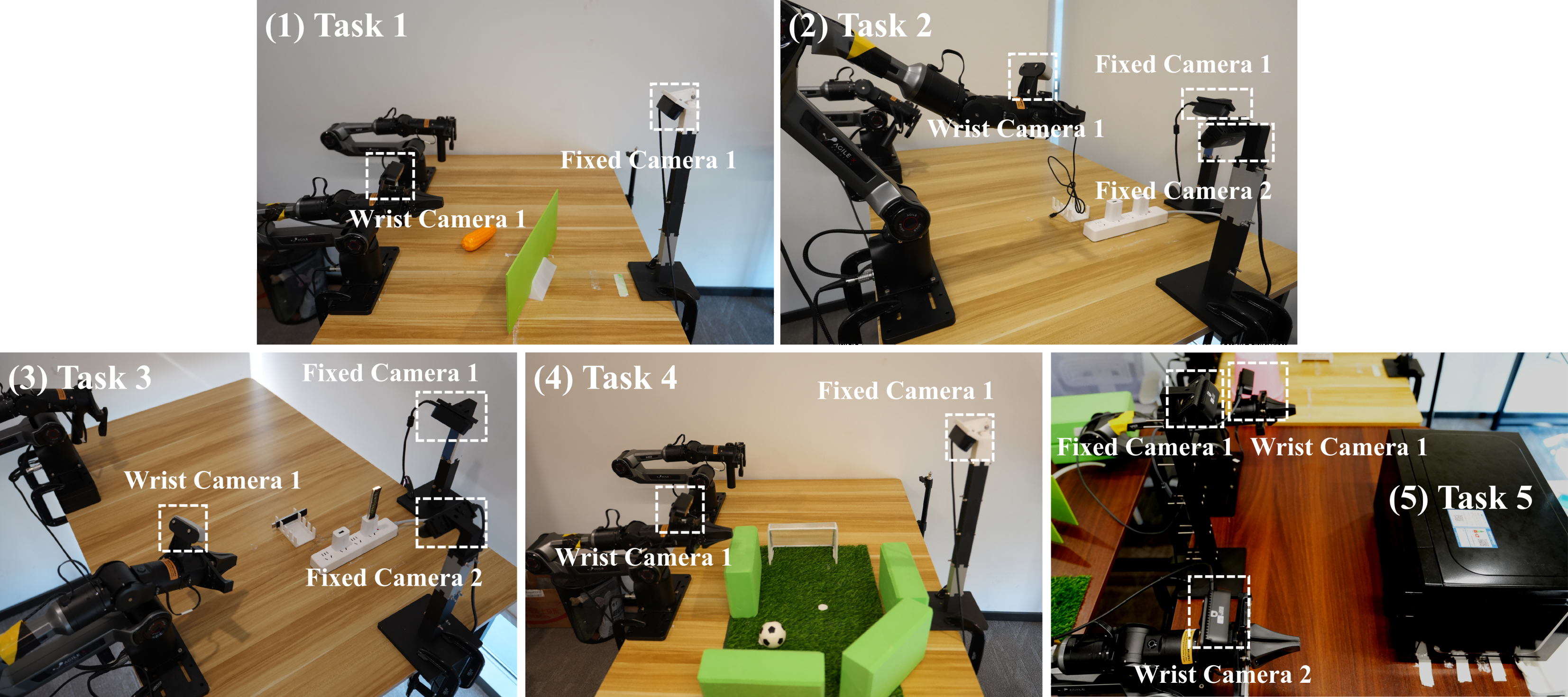}
  \caption{\textbf{Real-world experimental setups and camera configurations.}
The five tasks are: (1) Barrier-Crossing Corn Placement,
(2) USB Insertion, (3) Dual-USB Insertion,
(4) Obstacle-Course Soccer, and (5) Bimanual Microwave Retrieval.
Each setup includes wrist-mounted and fixed-view cameras,
with their locations highlighted by dashed boxes.
All cameras are Orbbec Gemini 335 RGB-D cameras.
For each task, observations from all cameras are used to train
the VLA and WAM models, whereas MAIF is trained using only
selected fixed-camera views.}
  \label{fig:experiment_details}
\end{figure}

\paragraph{Tasks.}
We evaluate four single-arm tasks and one bimanual task, as shown
in Fig.~\ref{fig:Experiment}.
\textbf{Task 1: Barrier-Crossing Corn Placement} requires lifting
an ear of corn over barriers of varying heights and placing it
into a bowl;
\textbf{Task 2: USB Insertion} requires aligning and inserting
a USB connector into its port;
\textbf{Task 3: Dual-USB Insertion} requires sequentially inserting
two USB connectors into their corresponding ports using a single arm;
\textbf{Task 4: Obstacle-Course Soccer} requires dribbling a ball
around obstacles before shooting it into a goal; and
\textbf{Task 5: Bimanual Microwave Retrieval} requires one arm
to open the microwave door and retrieve a box while the other
pushes the door aside.
Together, these tasks require metric height perception,
fine-grained spatial alignment, obstacle-aware planning,
and accurate depth perception in confined spaces.
The specific settings for the five tasks are shown in Fig.~\ref{fig:experiment_details}.

\begin{figure}[!h]
  \centering
  \includegraphics[width=1.0\linewidth]{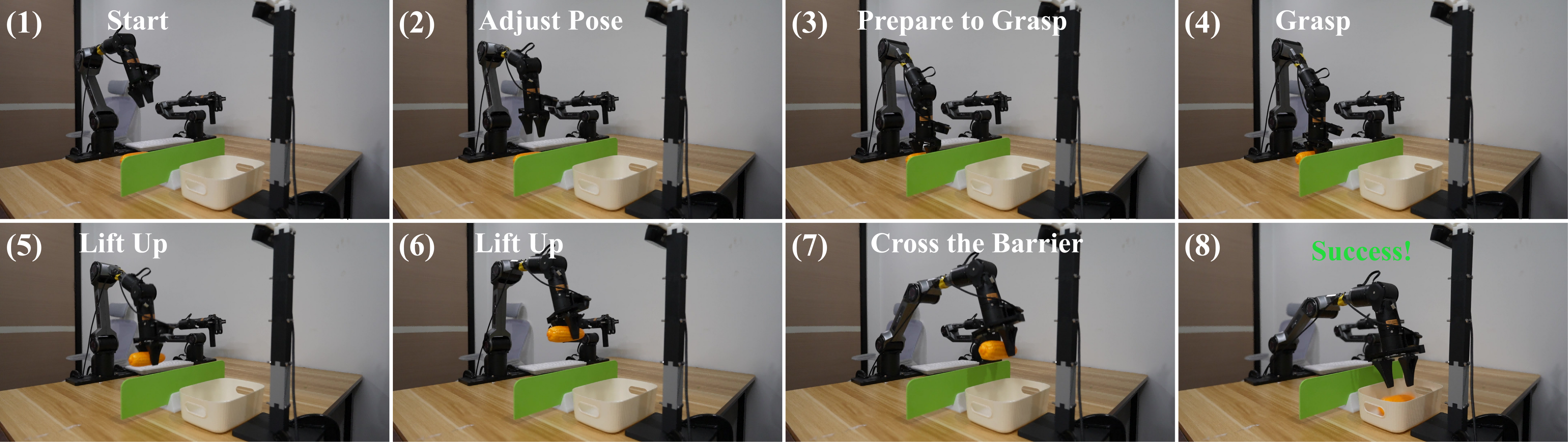}
   \caption{\textbf{Successful execution of Task 1: Barrier-Crossing Corn Placement.}
    The sequence shows our framework operating with the lowest barrier.
    The robot grasps the corn (1--4), lifts it above the barrier (5--6),
    moves it across (7), and places it into the bowl (8).
    The task requires perceiving the barrier height and controlling
    the lifting trajectory to clear the obstacle.} 
  \label{fig:corn_appendix}
\end{figure}

\begin{figure}[!h]
  \centering
  \includegraphics[width=1.0\linewidth]{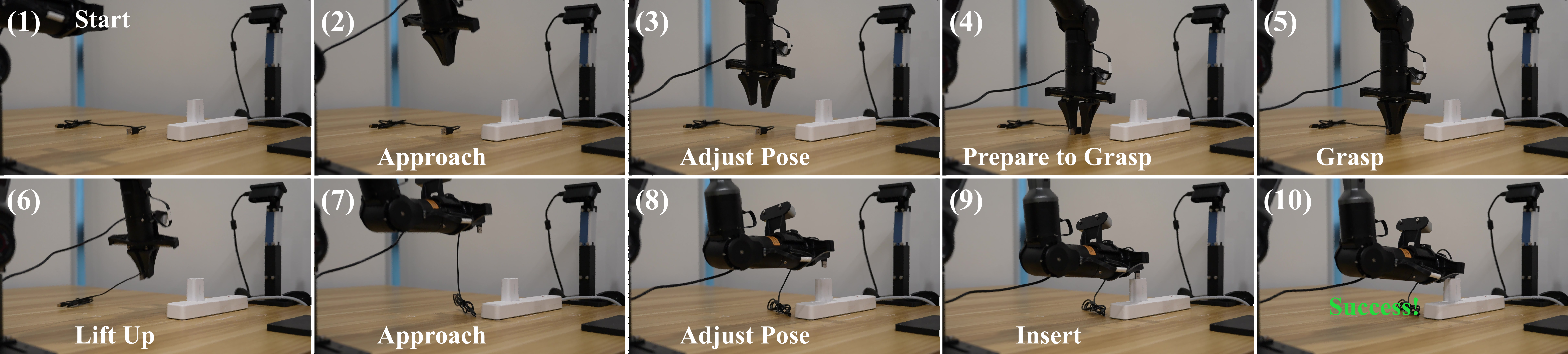}
  \caption{\textbf{Successful execution of Task 2: USB Insertion.}
    Our framework uses a single arm to approach and grasp the USB connector
    (1--5), lift and align it with the port (6--8), and complete the insertion
    (9--10). The task requires precise positional and rotational alignment
    to guide the connector into the narrow port.}
  \label{fig:singleusb_appendix}
\end{figure}

\begin{figure}[!h]
  \centering
  \includegraphics[width=1.0\linewidth]{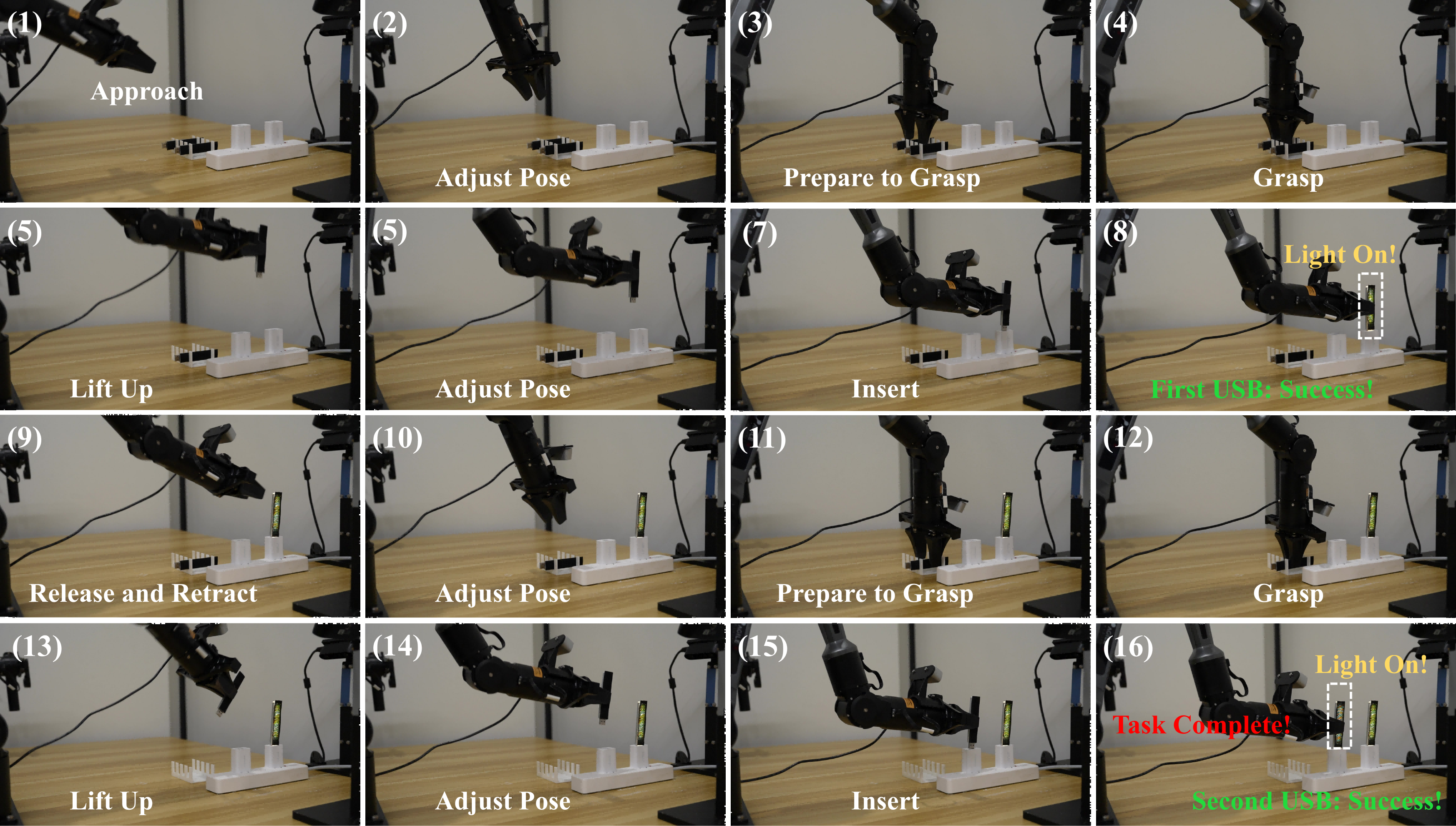}
  \caption{\textbf{Successful long-horizon execution of Task 3: Dual-USB Insertion.}
    Our framework uses a single arm to grasp, align, and insert the front
    USB connector first (1--8), followed by the rear connector (9--16).
    Each insertion is considered successful only when the corresponding
    indicator lights up, and the task is complete only after both connectors
    are inserted and both indicators are on.}
  \label{fig:dualusb_appendix}
\end{figure}

\begin{figure}[!h]
  \centering
  \includegraphics[width=1.0\linewidth]{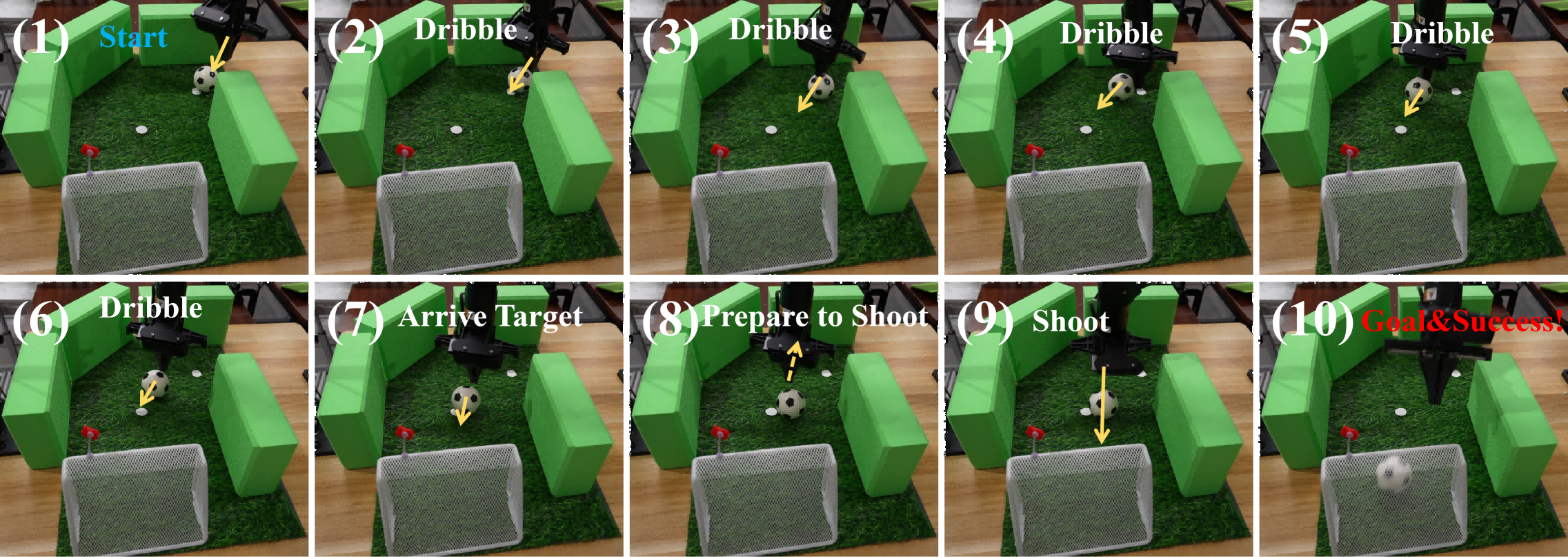}
    \caption{\textbf{Successful execution of Task 4: Obstacle-Course Soccer.}
    Our framework dribbles the ball around obstacles to the shooting position
    (1--7), draws back the gripper to prepare the shot (8), and shoots the ball
    into the goal (9--10).
    The task requires precise dribbling and obstacle avoidance in confined
    spaces; any contact between the gripper and the surrounding walls is
    counted as a failure.}
  \label{fig:soccer_appendix}
\end{figure}

\begin{figure}[!h]
  \centering
  \includegraphics[width=1.0\linewidth]{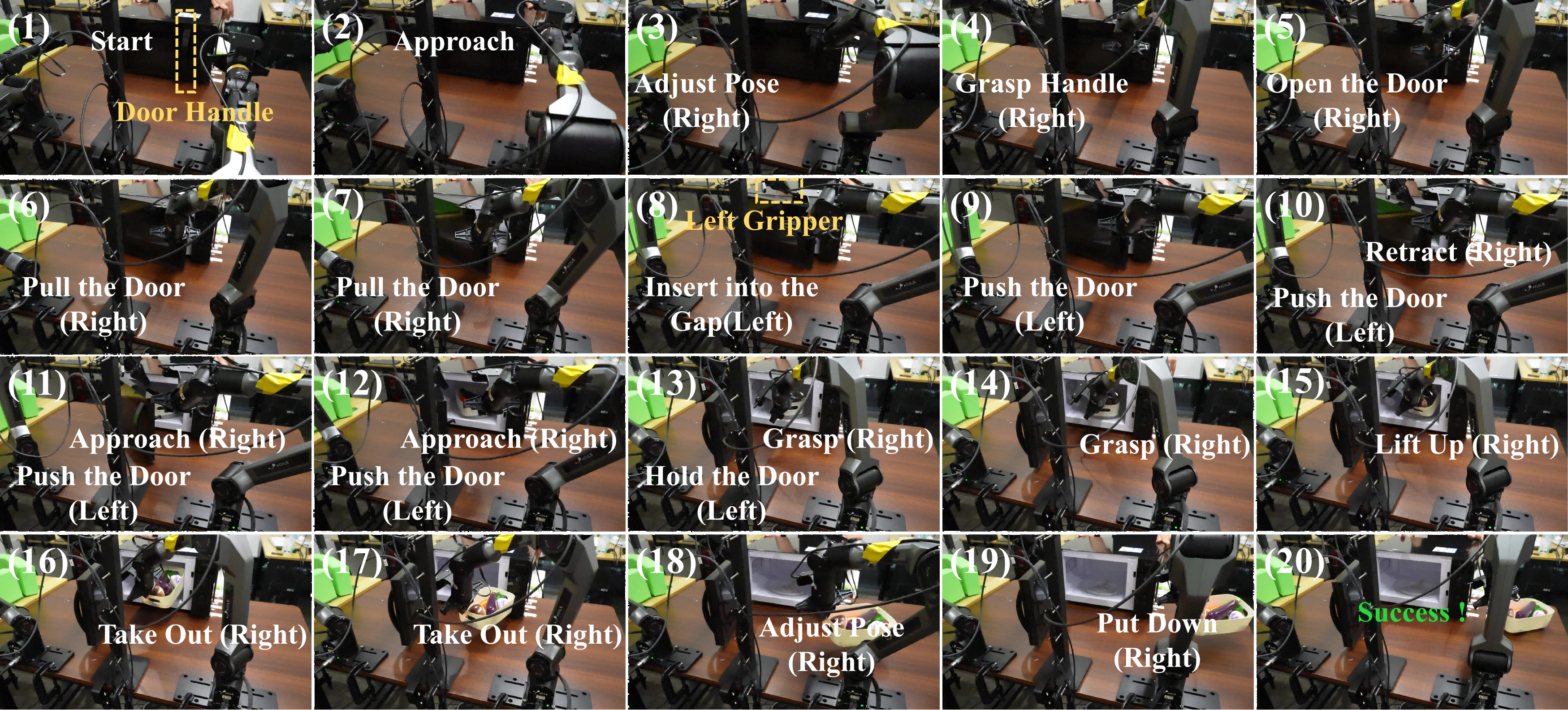}
    \caption{\textbf{Successful execution of Task 5: Bimanual Microwave Retrieval.}
    The right arm initially opens the door, with its opening range limited
    by the grasping pose. The left gripper then pushes the door farther open
    and holds it while the right arm retrieves the vegetable box and places
    it on the table. The task requires coordinated bimanual manipulation
    and stable transport; any spillage of vegetables is counted as a failure.}
  \label{fig:microwave_appendix}
\end{figure}

\paragraph{Visual observations.}
Both $\pi_{0.5}$ and Fast-WAM use the task-specific RGB camera
configurations summarized in Table~\ref{tab:task_camera_settings}.
For Task 5, one wrist-mounted camera is used on each arm.
All views are resized from $480 \times 640$ to $224 \times 224$
and use identity normalization. When training the Fast-WAM and $\pi_{0.5}$ baselines, we use RGB
observations from all cameras available for each task.
For MAIF training, point clouds are obtained only from the designated
fixed camera for each task.

\paragraph{Task Definition and Process Visualization.}
Figs~\ref{fig:corn_appendix}, \ref{fig:singleusb_appendix},
\ref{fig:dualusb_appendix}, \ref{fig:soccer_appendix},
and~\ref{fig:microwave_appendix} present detailed successful executions
of our framework on all five real-world tasks.

\begin{table}[h]
    \centering
    \caption{\textbf{Camera configurations} shared by both baselines.
    Task numbers correspond to the descriptions above.}
    \label{tab:task_camera_settings}
    \small
    \renewcommand{\arraystretch}{1.1}
    \setlength{\tabcolsep}{8pt}
    \begin{tabular}{@{}lccc@{}}
        \toprule
        \textbf{Tasks}
            & \textbf{Setup}
            & \textbf{Fixed cameras}
            & \textbf{Wrist cameras} \\
        \midrule
        Tasks 1 and 4 & Single-arm & 1 & 1 \\
        Tasks 2 and 3 & Single-arm & 2 & 1 \\
        Task 5        & Bimanual   & 1 & 2 \\
        \bottomrule
    \end{tabular}
\end{table}

\begin{table}[h]
    \centering
    \caption{\textbf{Point cloud sources} for MAIF training in each task.}
    \label{tab:maif_point_cloud_sources}
    \small
    \resizebox{\linewidth}{!}{%
        \begin{tabular}{@{}lccccc@{}}
            \toprule
            \textbf{Task}
                & \textbf{Task 1} & \textbf{Task 2}
                & \textbf{Task 3} & \textbf{Task 4}
                & \textbf{Task 5} \\
            \midrule
            Point cloud source
                & Fixed Camera 1 & Fixed Camera 2
                & Fixed Camera 2 & Fixed Camera 1
                & Fixed Camera 1 \\
            \bottomrule
        \end{tabular}%
    }
\end{table}

\begin{table}[h]
    \centering
    \caption{\textbf{Number of expert episodes} per task in real-world experiments.}
    \label{tab:realworld-episode-num}
    \small
    \resizebox{0.6\linewidth}{!}{%
        \begin{tabular}{@{}lccccc@{}}
            \toprule
            \textbf{Task}
                & \textbf{Task 1} & \textbf{Task 2}
                & \textbf{Task 3} & \textbf{Task 4}
                & \textbf{Task 5} \\
            \midrule
            Episode Number
                & 100 & 100
                & 100 & 100
                & 200 \\
            \bottomrule
        \end{tabular}%
    }
\end{table}

\paragraph{Baselines Setup.}
Both baselines use one proprioceptive observation step.
The $\pi_{0.5}$ model interface pads states and actions to 32-D.
Fast-WAM uses 7-D states and actions for Tasks 1--4 and
14-D states and actions for Task 5.
The language sequence lengths are 200 tokens for $\pi_{0.5}$
and 128 tokens for Fast-WAM. Both baselines predict 50-step action chunks and execute all using
10 denoising steps at inference. For each baseline and task, we use the terminal training checkpoint,
fixed a priori without held-out model selection.
We conduct 100 evaluation trials per task for each baseline.

\paragraph{Our Framework.}
Our framework follows a two-stage training procedure.
In Stage~1, professional annotators provide pose annotations
on 2D images. These annotations are lifted to 3D using the
corresponding depth measurements from RGB-D cameras to
provide supervision for ICT. Stage~1 uses the same training hyperparameters as the corresponding baseline.
In Stage~2, MAIF is trained using point clouds from designated
fixed cameras, with the source for each task summarized in
Table~\ref{tab:maif_point_cloud_sources}. For Tasks 2 and 3, which have multiple fixed cameras, we select the
camera whose placement provides a suitable sensing distance for accurate
depth acquisition with fewer missing regions, while covering the geometry
of the key manipulation area.

\paragraph{Data Collection.}
All real-world demonstration data are collected via teleoperation.
The amount of training data used for each task is summarized in Table~\ref{tab:realworld-episode-num}.

\subsection{Real-World Training Details}
\label{app:real_world_training_details}

\paragraph{Shared training settings.}
Both baselines are trained on 8 H200 GPUs using AdamW with
$\beta_1 = 0.9$, $\beta_2 = 0.95$, $\epsilon = 10^{-8}$,
and weight decay $0.01$.
We use gradient clipping at $1.0$ and no gradient accumulation.
Data augmentation, LoRA/PEFT, ACP, and AWR are disabled.
Model-specific training configurations are summarized in
Table~\ref{tab:training_settings}. The training configuration for ICT in Stage~1 is aligned with that of the corresponding baseline.

\paragraph{Fast-WAM preprocessing.}
For Fast-WAM, the video context consists of 51 time steps
subsampled into 9 video frames.
VAE latents and text embeddings are precomputed offline to accelerate training.

\begin{table}[h]
    \centering
    \caption{\textbf{Model-specific training configurations} of
    $\pi_{0.5}$ and Fast-WAM for the real-world experiments.
    Shared training settings are described in the text.}
    \label{tab:training_settings}
    \label{tab:fastwam_training_settings}
    \footnotesize
    \renewcommand{\arraystretch}{1.05}
    \setlength{\tabcolsep}{4pt}
    \begin{tabularx}{\linewidth}{@{}p{0.21\linewidth}XX@{}}
        \toprule
        \textbf{Setting}
            & \textbf{$\pi_{0.5}$}
            & \textbf{Fast-WAM} \\
        \midrule
        Initialization
            & \texttt{lerobot/pi05\_base}
            & \texttt{lerobot/fast-wam\_base};
              video expert:
              \texttt{Wan-AI/Wan2.2-TI2V-5B} \\
        Architecture
            & PaliGemma (\texttt{gemma\_2b}) with a
              \texttt{gemma\_300m} action expert
            & Layer-wise mixture of a Wan2.2 video DiT
              and an action DiT \\
        Fine-tuning
            & All parameters, including the vision encoder
              and VLM
            & All parameters of both DiTs and the
              proprioceptive encoder \\
        Frozen modules
            & None
            & Wan VAE and UMT5 text encoder \\
        Precision
            & FP32; AMP disabled
            & BF16 training; FP32 loss computation \\
        State/action normalization
            & Quantile
            & Min--max scaling to $[-1,1]$ \\
        \midrule
        Objective
            & Conditional flow-matching velocity regression
              with MSE loss
            & Joint video--action conditional flow matching
              with equally weighted MSE losses \\
        Time sampling
            & $u \sim \mathrm{Beta}(1.5,1.0)$;
              \newline
              $t = 0.999u + 0.001$
            & $u \sim \mathcal{U}(0,1)$;
              \newline
              $\sigma = 5u/(1+4u)$;
              \newline
              $t = 1000\sigma$ \\
        Peak / minimum LR
            & $2.5 \times 10^{-5}$ /
              $2.5 \times 10^{-6}$
            & $10^{-4}$ / $10^{-6}$ \\
        LR schedule
            & 1,000-step warmup;
              30,000-step cosine decay
            & 5\% linear warmup followed by cosine decay \\
        Batch size
            & 8 per GPU;
              global batch size 64
            & Two-view tasks: 32 per GPU, global 256;
              three-view tasks: 24 per GPU, global 192 \\
        Optimizer steps
            & 50,000 for all tasks
            & 50,000 for Tasks 1--4;
              80,000 for Task 5 \\
        Random seed
            & 1000
            & 42 \\
        \midrule
        Distributed training
            & 8-process DDP
            & 8-process ZeRO-1 \\
        Gradient checkpoint
            & Enabled
            & Disabled \\
        \texttt{torch.compile}
            & Enabled with \texttt{max-autotune}
            & Disabled \\
        Data loading
            & 4 workers per rank
            & 4 persistent workers per rank;
              prefetch factor 4; pinned memory;
              asynchronous CPU--GPU transfer \\
        \bottomrule
    \end{tabularx}
\end{table}

\paragraph{MAIF training settings.}
For both $\pi_{0.5}$ and Fast-WAM, MAIF is initialized on top of the
corresponding task-specific real-world policy checkpoint.
The complete backbone policy and the Sonata point encoder remain frozen,
and only the MAIF parameters are optimized.
Training uses only successful demonstrations.
Calibrated depth observations are back-projected into the robot base frame
and uniformly sampled into 8,192 metric points.
Each point is represented by its position and surface normal, while the original coordinates in meters are preserved without
per-scene centering or scale normalization. The detailed training settings for MAIF in real-world experiments are provided in Table~\ref{tab:maif_realworld_training}.

Both implementations are trained on 8 H200 GPUs for 5,000 optimizer steps
with a per-GPU batch size of 32, giving a global batch size of 256.
We use AdamW with $\beta_1=0.9$, $\beta_2=0.95$,
$\epsilon=10^{-8}$, weight decay $10^{-4}$, and gradient clipping at
$1.0$.
The learning rate is warmed up for 500 steps to $10^{-4}$ and then
decayed using a cosine schedule.
No gradient accumulation, LoRA/PEFT, or backbone fine-tuning is used.
All reported real-world results use the checkpoint at step 5,000.


\begin{table}[h]
    \centering
    \caption{\textbf{MAIF training configurations for the real-world experiments.}
    The task-specific policy and Sonata encoder are frozen in both implementations.}
    \label{tab:maif_realworld_training}
    \footnotesize
    \renewcommand{\arraystretch}{1.05}
    \setlength{\tabcolsep}{4pt}

    \newcommand{\maifshared}[1]{%
        \multicolumn{2}{%
            >{\raggedright\arraybackslash}
            p{\dimexpr0.75\linewidth-2\tabcolsep\relax}@{}%
        }{#1}%
    }

    \begin{tabularx}{\linewidth}{
        @{}
        >{\raggedright\arraybackslash}p{0.25\linewidth}
        >{\raggedright\arraybackslash}X
        >{\raggedright\arraybackslash}X
        @{}
    }
        \toprule
        \textbf{Setting}
            & \textbf{$\pi_{0.5}$ + MAIF}
            & \textbf{FastWAM + MAIF} \\
        \midrule
        Initialization
            & Task-specific $\pi_{0.5}$ checkpoint
            & Task-specific FastWAM checkpoint \\
        Trainable modules
            & MAIF only
            & MAIF only \\
        Frozen modules
            & PaliGemma, action expert, and Sonata
            & Video DiT, action DiT, encoders, and Sonata \\
        Action--metric bridge
            & Quantile de-normalization followed by fixed robot FK
            & Min--max de-normalization followed by fixed robot FK \\
        Action horizon
            & 50
            & 50 \\
        \midrule
        Metric geometry
            & \maifshared{
                8,192 base-frame points in meters;
                frozen Sonata with a finest resolution of
                $2\,\mathrm{cm}$
            } \\
        Point features
            & \maifshared{
                XYZ + normal
            } \\
        MAIF architecture
            & \maifshared{
                Hidden size 1,024; 16 heads;
                3 geometry cross-attention, 10 temporal,
                and 1 bimanual interaction layer
            } \\
        Precision
            & \maifshared{
                BF16 network computation;
                FP32 metric decoding and loss computation
            } \\
        \midrule
        Batch size
            & \maifshared{
                32 per GPU; global batch size 256
            } \\
        Optimizer steps
            & \maifshared{5,000} \\
        LR schedule
            & \maifshared{
                500-step warmup to $10^{-4}$; cosine decay
            } \\
        Optimizer
            & \maifshared{
                AdamW, weight decay $10^{-4}$,
                gradient clipping $1.0$
            } \\
        Gradient accumulation
            & \maifshared{None} \\
        Distributed training
            & 8-process DDP
            & 8-process ZeRO-1 \\
        Random seed
            & \maifshared{42} \\
        Evaluation checkpoint
            & \maifshared{Step 5,000} \\
        \bottomrule
    \end{tabularx}
\end{table}

\subsection{Real World Experiment Visualization}

\begin{figure}[!h]
  \centering
  \includegraphics[width=1.0\linewidth]{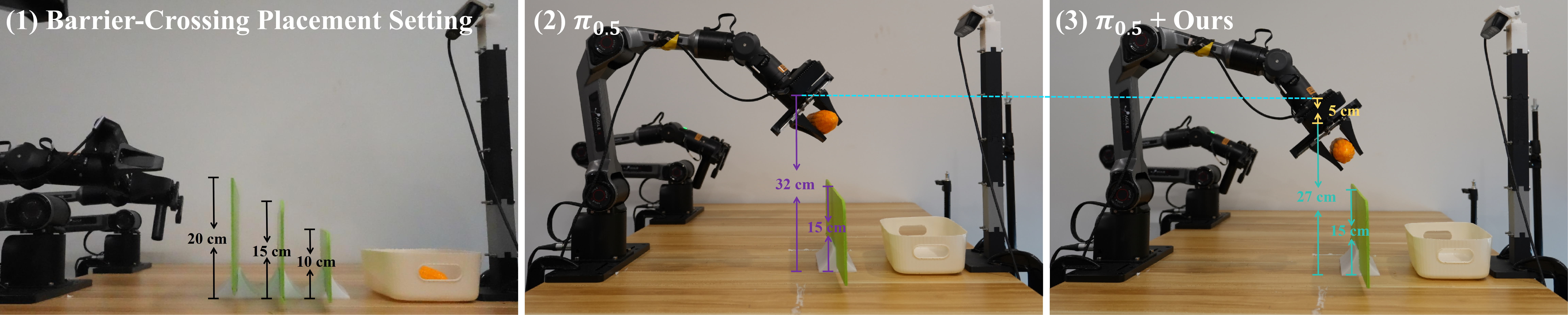}
  \caption{\textbf{Height-adaptive trajectories in barrier-crossing placement.}
    (1) Task setup with three barrier heights: $10$, $15$, and $20\,\mathrm{cm}$.
    (2) $\pi_{0.5}$ + Ours crosses the same barrier at $27\,\mathrm{cm}$, producing a lower trajectory better matched to the actual obstacle height. This height-adaptive behavior illustrates improved metric spatial awareness and avoids unnecessary lifting.
    (2) When crossing the $15\,\mathrm{cm}$ barrier, $\pi_{0.5}$ lifts the corn to $32\,\mathrm{cm}$.
}
  \label{fig:Corn}
\end{figure}

\begin{table}[!t]
    \centering
    \caption{Statistics on the height of crossing the barrier.}
    \label{tab:corn_cross_height}
    \small
    \resizebox{0.8\linewidth}{!}{%
        \begin{tabular}{@{}lcccccc@{}}
            \toprule
            \textbf{Method \& Height (mm)}
                & \textbf{Test 1} & \textbf{Test 2}
                & \textbf{Test 3} & \textbf{Test 4}
                & \textbf{Test 5} & \textbf{Average $\downarrow$} \\
            \midrule
            $\pi_{0.5}$
                & 301.617 & 295.685
                & 318.257 & 325.743
                & 294.452 & 307.151 \\
            $\pi_{0.5}$ + Ours
                & 270.125 & 284.773
                & 291.781 & 301.127
                & 265.301 & \textbf{282.621} \\
            \bottomrule
        \end{tabular}%
    }
\end{table}

Fig.~\ref{fig:Corn} visualizes Task~1,
Barrier-Crossing Corn Placement. During data collection, we used
barriers of three heights (10, 15, and 20\,cm) and collected
demonstrations with crossing heights adapted to each barrier,
as shown in panel~(1). Although both $\pi_{0.5}$ and $\pi_{0.5}$ + Ours achieve
100\% success on this task, their crossing trajectories differ.
For the 15\,cm barrier, our framework crosses at 27\,cm,
whereas $\pi_{0.5}$ lifts the corn to 32\,cm, resulting in
unnecessary vertical motion.

Table~\ref{tab:corn_cross_height} reports the mean peak end-effector height over five trials
of corn placement across the 15\,cm barrier.
Our framework achieves a mean peak height of 282.621\,mm, compared
with 307.151\,mm for the $\pi_{0.5}$ baseline.
This lower lifting height suggests that metric height awareness
helps our framework learn a trajectory better matched to the
medium-height barrier, reducing unnecessary lifting.

\subsection{Limitations}

Our framework has three main limitations.
First, its performance depends on point cloud quality:
depth noise from RGB-D cameras can produce noisy and discontinuous
scans, potentially affecting downstream performance.
Future work will explore more robust alternatives to point cloud
representations.
Second, our interaction mechanism requires accurate calibration
and alignment between coordinate frames, which complicates
the hardware setup.
Third, our experiments are limited to manipulation with grippers.
We plan to extend the evaluation to dexterous hands in future work.

\end{document}